\documentclass[10pt, journal]{IEEEtran}

\usepackage{times}
\usepackage{multicol}
\usepackage[bookmarks=true]{hyperref}
\usepackage{xcolor}
\usepackage{hyperref}
\usepackage{amsmath, amssymb}
\usepackage{amsfonts}
\usepackage{graphicx}
\usepackage{siunitx}
\usepackage{standalone}
\usepackage{booktabs}
\usepackage[ruled,vlined,linesnumbered]{algorithm2e}
\usepackage{mdframed}
\usepackage{fancyvrb,multirow}
\usepackage{soul}
\usepackage{dsfont,mathabx}
\usepackage{array, booktabs}
\usepackage{subfigure}
\usepackage{amsthm}
\usepackage{threeparttable}

\newtheorem{theorem}{Theorem}

\newtheorem{lemma}[theorem]{Lemma}

\theoremstyle{definition}
\newtheorem{definition}{Definition}
\newtheorem{remark}{Remark}
\newtheorem{example}{Example}
\newtheorem{problem}{Problem}

\title{Hierarchical Motion Planning under Probabilistic Temporal Tasks
  and Safe-Return Constraints}

\author{Meng Guo$^1$, Tianjun Liao$^2$, Junjie Wang$^1$ and Zhongkui Li$^1$ 
  \thanks{The authors are with $^1$the State Key Laboratory
    for Turbulence and Complex Systems,
    Department of Mechanics and Engineering Science,
    College of Engineering, Peking University, Beijing, China;
    $^2$the Academy of Military Sciences, Beijing, China.
    This work was supported by the National Natural Science Foundation
    of China under grants 62203017, T2121002, U2241214;
    by the Ministry of Education under grant 2021ZYA05004;
    and by Beijing Natural Science Foundation under grant JQ20025.
    Corresponding author: \texttt{zhongkli@pku.edu.cn}.
}
}

\begin{document}

\maketitle

\begin{abstract}
Safety is crucial for robotic missions within an uncertain environment.
Common safety requirements such as collision avoidance are only state-dependent,
which can be restrictive for complex missions.
In this work, we address a more general formulation as safe-return constraints,
which require the existence of a return-policy to drive the system back to a set of safe states with high probability.
The robot motion is modeled as a Markov Decision Process (MDP) with probabilistic labels,
which can be highly non-ergodic.
The robotic task is specified as Linear Temporal Logic (LTL) formulas over these labels,
such as surveillance and transportation.
We first provide theoretical guarantees on the re-formulation of such safe-return constraints,
and a baseline solution based on computing two complete product automata.
Furthermore, to tackle the computational complexity,
we propose a hierarchical planning algorithm that combines the feature-based symbolic and temporal abstraction with constrained optimization.
It synthesizes simultaneously two dependent motion policies:
the outbound policy minimizes the overall cost of satisfying the task with a high probability,
while the return policy ensures the safe-return constraints.
The problem formulation is versatile regarding the robot model, task specifications and safety constraints.
The proposed hierarchical algorithm is more efficient and can solve much larger problems than the baseline solution, with only a slight loss of optimality.
Numerical validations include simulations and hardware experiments
of a search-and-rescue mission and a planetary exploration mission over various system sizes.
\end{abstract}


\section{Introduction}\label{sec:introduction}
{Autonomous mobile robots are often deployed in uncertain and unsafe environments
  that are otherwise risky for humans. For instance,}
an autonomous ground vehicle (AGV) is deployed in an office for a
search-and-rescue mission after disaster:
to search for injured victims and bring them to medical stations,
to shut down certain machines in different rooms,
and to report fire hazards or gas leakage;
also an autonomous rover is deployed in a rough terrain for a planetary exploration mission:
to gather specimens from areas of interest,
to assemble them into containers and store in the storage area;
and to charge often.
Different from simple navigation tasks,
such tasks require not only planning on the task-level regarding which areas to visit and actions to perform there,
but also planning on the motion-level regarding how to reach a desired area.
In some cases, these tasks are long-term meaning that they should be repeated infinitely often.
{Furthermore, even though the approximate structure of the environment is known,
its exact features within the structure can only be estimated.}
This has two direct consequences:
first, the robot movement within the environment become uncertain,
e.g., drifting due to different terrains, and blockage due to debris;
second, the properties relevant to the task become uncertain,
e.g., which areas contains human victims, and which charging station is functional.
In other words, uncertainty in the environment model can cause non-determinism both in the task planning and motion execution.
A formal consideration of this uncertainty when planning for general complex tasks remains challenging.
Some recent work addresses this issue with model-based optimization~\cite{guo2018probabilistic,belta2017formal,ding2014ltl} or
data-driven online learning~\cite{hasanbeig2019reinforcement, bozkurt2020control}.
However, the computational complexity remains to be the bottleneck and most validations are performed on systems with small number of states.

On the other hand, safety is an indispensable aspect to consider when deploying robotic systems,
especially so in an uncertain and potentially unsafe environment.
Depending on the application, robotic safety can be defined in various ways.
Safety in motion planning~\cite{lavalle2006planning} is commonly defined as the avoidance of a subset of the state space, namely the \emph{unsafe} regions.
This has been made more general with temporal task specifications,
e.g., always stop in front of human, infinitely often charge itself, and always travel at the right lane, see~\cite{guo2018probabilistic, hasanbeig2019reinforcement, laurenti2020formal, guo2015multi, schillinger2018simultaneous}.
These safety rules are useful and often directly imposed as an additional requirement of the overall task.
However,
there are certain types of safety requirements that can \emph{not} be expressed in this way.
Safe-return constraints require that the robot should be able to return to a set of safety states whenever requested during the mission.
Such constraints are particularly important for non-ergodic systems where not all states are reachable for any other states.
Considering the same example of search-and-rescue missions,
the robot should avoid one-way doors, stairs and debris where it may get trapped in some rooms;
during the planetary exploration mission,
the robot should avoid steep cliffs and valleys where it can easily descend but hard to ascend back.
In other words, such safe-return constraints are less state-dependent but more \emph{policy-dependent},
thus more general in terms of expressiveness and practical relevance.
More importantly, these constraints are often {conflicting} with the long-term task goals,
e.g., the task specifies to search and rescue victims at all time, while the safe-return constraints require to return and stay at the base.
Consequently, they can not be added directly in the task specification,
instead a separate return policy should be synthesized to fulfill the constraints and further restrict the task execution.
To the best of our knowledge, such safe-return constraints have been mainly addressed
in the reinforcement learning community~\cite{moldovan2012safe,turchetta2016safe}
to ensure safety during the learning process.
They have not been studied along with the complex temporal tasks as in this work.

In this work, we address the motion planning problem over MDPs with probabilistic labels,
where the tasks are given as Linear Temporal Logic (LTL) formulas.
Furthermore, we take into account a general safety requirement as safe-return constraints,
which demands the existence of a return-policy to drive the system back to home states with a high probability.
It is first shown that the safe-return constraints can be re-formulated as accumulated rewards
given the value function in the associated safety automaton.
Then, the baseline solution is proposed by solving two coupled linear programs in the respective product automaton.
However, this solution quickly becomes intractable when the system size increases.
{To overcome this bottleneck, a hierarchical and approximate planning algorithm based on the combination of symbolic and temporal abstractions with constrained optimization is proposed.}
Two abstracted semi-MDPs and the associated motion policies as \emph{temporal options} are constructed simultaneously for the regions whose features are relevant to the task and safety specifications.
Afterwards, two high-level task policies are synthesized in sequence within the respective product automata between the semi-MDPs and task automata.
The return policy ensures that the safe-return constraints are fulfilled, and the outbound policy minimizes the overall cost to satisfy the task.
{It is shown that this hierarchical solution is more efficient and can solve systems of much larger sizes,
compared with the alternative and baseline solutions, with only a slight reduction of optimality}.
The results are validated rigorously via simulations and hardware experiments of various robotic missions
with different system sizes.

The main contribution lies in three aspects:
(i) {the formulation of a new planning problem for MDPs with labelling features,
  where both the tasks and safe-return constraints are given as complex LTL formulas};
(ii) the theoretical analyses that ensure the correctness of the problem re-formulation;
(iii) the hierarchical planning algorithm that synthesizes simultaneously and efficiently the outbound policy for task satisfaction and the return policy for safety constraints.

The rest of the paper is organized as follows: Sec.~\ref{sec:prelims}  introduces some preliminaries of LTL.
The problem formulation is given in Sec.~\ref{sec:problem-formulate}.
Sec.~\ref{sec:analysis} provides the theoretical analyses to re-formulate the problem and therefore the baseline solution.
The hierarchical planning framework is presented in Sec.~\ref{sec:solution}.
Simulation and experiment results are shown in Sec.~\ref{sec:case}.
Finally, Sec.~\ref{sec:future} concludes with future work.

\section{Related Work}\label{sec:related-work}

\subsection{MDPs with Temporal Tasks}\label{subsec:mdp-ltl}
MDPs provide a powerful model for robots acting in an uncertain environment.
Uncertainty arises in various aspects of the system such as the properties of the workspace and the outcome of an action~\cite{puterman2014markov}.
A common objective is to reach a set of terminal states while maximizing {expected cumulative reward}.
When the underlying MDP is fully-known,
a variety of algorithms can be applied to find the optimal acting policy for the robot,
such as dynamic programming~\cite{bellman1966dynamic, bertsekas1995neuro}, value iteration and policy iteration~\cite{puterman2014markov}.
The resulting policy maps the current state to the set of optimal actions,
deterministically or stochastically.
Otherwise, if the underlying MDP is only partially-known,
{online learning algorithms~\cite{sutton2018reinforcement} can be used.}
Furthermore, instead of the simple reachability task,
there have been many efforts to address the same problem,
rather to satisfy high-level temporal tasks specified in various formal languages,
such as Probabilistic Computation Tree Logic (PCTL) in~\cite{lahijanian2015formal}
and Linear Temporal Logics (LTL) in~\cite{guo2018probabilistic, belta2017formal, ding2014optimal,  forejt2011quantitative}.
Such languages are intuitive and expressive when specifying complex control tasks,
such as surveillance, transportation and emergency response, see~\cite{guo2015multi, tumova2016multi, kantaros2018control, schillinger2018simultaneous}.
{Different cost optimizations are considered along with the task satisfiability,
such as maximum reachability for constrained MDPs in~\cite{ding2014optimal,forejt2011quantitative},
  the minimal bottleneck cost in~\cite{smith2011optimal},
  and pareto curves under multiple temporal objectives in~\cite{Forejt12Pareto}.}
Verification toolboxes are provided in~\cite{etessami2007multi, kwiatkowska2011prism} for certain task formats.
Moreover, robust control policies for a temporal task are studied when the underlying MDP is uncertain,
for instance,~\cite{ding2014ltl} maximize the accumulated time-varying rewards, ~\cite{wolff2012robust} maximizes the satisfiability under uncertain transition measures,
and~our previous work~\cite{guo2018probabilistic} allows for infeasible tasks to be only partially satisfied by the MDP.
Lastly, some recent work builds upon methods from reinforcement learning to combine data-driven approach with the aforementioned model-based planning methods.
For example, the work~\cite{hasanbeig2019reinforcement}  constructs the product automaton on-the-fly while interacting with the environment.
The work~\cite{bozkurt2020control} instead proposes a reward shaping method to maximize the task satisfiability without directly learning the underlying transition model,
while~\cite{wang2015temporal} relies on the actor-critic methods to approximate over large set of state-action pairs.
Due to the exponential complexity and data inefficiency,
most of the work above is validated over case studies of limited sizes.
{In this work, we assume the model as a known MDP with labelling features,
to focus instead on the consideration of safe-return constraints
and more importantly, feature-based symbolic and temporal abstraction methods
to reduce the computational complexity.}

\subsection{Safety in Planning}\label{subsec:safety}
Safety in motion planning~\cite{lavalle2006planning} is commonly defined as the avoidance of a subset of the state space, namely the {unsafe} regions.
In the literature for planning over temporal tasks,
it can be more general and specified directly as an additional requirement in the task specification,
e.g., ``always avoid obstacles'', ``infinitely often visit charging station'' and ``always stop in front of human'', see~\cite{guo2018probabilistic, hasanbeig2019reinforcement, laurenti2020formal, guo2015multi, schillinger2018simultaneous}.
Consequently, they are treated similarly as other performance-related tasks.
The work in~\cite{tumova2013least, vasile2017minimum} specifies a variety of safety rules as sub-task formulas and synthesizes a minimum-violating policy for these rules.
Our earlier work~\cite{guo2015multi} proposes to separate the performance and safety requirements as soft and hard specification, respectively.
The hard specifications have to be satisfied at all time and soft specifications can be improved gradually during exploration.
In contrast, the safety constraints considered in this work as \emph{safe-return} constraints,
which require that the robot should be able to return to a set of safety states anytime when requested during the mission.
They are particularly important for non-ergodic systems,
and significantly different from the aforementioned definitions:
(i) whether a region is unsafe depends on the existence of a return policy, rather on the region itself;
(ii) two dependent polices, outbound policy and return policy, should be constructed,
whereas traditionally only one policy is needed for both the task and safety specifications~\cite{guo2018probabilistic, laurenti2020formal, guo2015multi, vasile2017minimum};
(iii) the safe-return  constraints are often {conflicting} with the task goal,
thus can not be added directly to the task specification;
(iv) traditional safety definitions mentioned above should be included in task specification,  rather than the safe-return constraints.
{Such safe-return constraints have been shown to be useful during reinforcement learning, e.g.,~\cite{moldovan2012safe,turchetta2016safe}.
However,  they have not been studied in the context of complex temporal tasks.
}

\subsection{Hierarchical Temporal and Symbolic Abstraction}\label{subsec:abstraction}
Uniform discretization of a continuous high-dimensional system often leads to complexity explosion,
thus is only applicable to toy cases.
{Temporal and symbolic abstraction techniques are proposed to address this complexity issue.
  The notion of \emph{options} as closed-loop policies for taking actions over a period of time is pioneered in~\cite{sutton1999between}, which allows for high-level planning over the semi-MDPs for extended planning horizons.
Feature-based or symbolic abstraction~\cite{tabuada2009verification} is another effective way to tackle the planning problem
sequentially at different levels of granularity.} For instance,
it is a common practice to plan first over regions of interest in the workspace given the temporal task, see~\cite{guo2015multi, ding2014optimal, tumova2016multi, kantaros2018control},
which is then executed by the low-level feedback motion controller to navigate among these regions.
Moreover,~\cite{meyer2019hierarchical, nilsson2014incremental} proposes an automated abstraction algorithm for a general nonlinear system to satisfy a LTL formula,
\cite{laurenti2020formal} constructs a finite abstraction model as uncertain MDPs for the class of switched diffusion systems.
Bi-simulation or approximation relations~\cite{haesaert2017verification} are widely used to represent the relation between the original system and the abstraction.
{In this work, the feature-based symbolic and temporal abstractions are combined and applied to
  both the safe-return constraints and the task specifications,
which are dependent and thus constructed simultaneously.}

\section{Preliminaries}\label{sec:prelims}

\subsection{Linear Temporal Logic (LTL)}\label{subsec:LTL}
The  ingredients of a Linear Temporal Logic (LTL) formula are a set of atomic propositions $AP$ and several Boolean and temporal operators. Atomic propositions are Boolean variables that can be either true or false.  A LTL formula is specified according to the  syntax~\cite{baier2008principles}:
$\varphi \triangleq \top \;|\; p  \;|\; \varphi_1 \wedge \varphi_2  \;|\; \neg \varphi  \;|\; \bigcirc \varphi  \;|\;  \varphi_1 \,\textsf{U}\, \varphi_2,$
where $\top\triangleq \texttt{True}$, $p \in AP$, $\bigcirc$ (\emph{next}), $\textsf{U}$ (\emph{until}) and $\bot\triangleq \neg \top$. For brevity, we omit the derivations of other operators like $\Box$ (\emph{always}), $\Diamond$ (\emph{eventually}), $\Rightarrow$ (\emph{implication}).
The semantics of LTL is defined over the set of infinite words over $2^{AP}$.
Intuitively, $p \in AP$ is satisfied on a word $w = w(1)w(2)...$ if it holds at $w(1)$,
i.e., if $p \in w(1)$. Formula $\bigcirc \varphi$ holds true if $\varphi$ is satisfied
on the word suffix that begins in the next position $w(2)$,
whereas $\varphi_1 \textsf{U} \varphi_2$ states that $\varphi_1$ has to be true until $\varphi_2$
becomes true. Finally, $\Diamond \varphi$ and $\Box \varphi$ are true if $\varphi$ holds on $w$
eventually and always, respectively.
Thus, given any word over~$AP$, it can be verified whether $w$ satisfies the formula,
denoted by~$w\models \varphi$.
The full semantics and syntax of LTL are omitted here due to limited space, see e.g.,~\cite{baier2008principles}.

\subsection{Deterministic Rabin Automaton~(DRA)}\label{subsec:dra}
The set of words that satisfy a LTL formula~$\varphi$ over $AP$ can be captured through a Deterministic Rabin Automaton~(DRA)~$\mathcal{A}_{\varphi}$~\cite{baier2008principles}, defined as~$\mathcal{A}_{\varphi}\triangleq (Q, \,2^{AP},\, \delta,\, q_0,\,\text{Acc}_{\mathcal{A}})$,
where $Q$ is a  set of states; {$2^{AP}$ is the alphabet}; $\delta\subseteq Q\times 2^{AP} \times {Q}$ is a transition relation; $q_0 \in Q$ is the initial state; and~$\text{Acc}_{\mathcal{A}} \subseteq 2^Q \times 2^Q$ is a set of accepting pairs, i.e., $\text{Acc}_{\mathcal{A}} = \{(H^1_{\mathcal{A}}, I^1_{\mathcal{A}}), (H^2_{\mathcal{A}}, I^2_{\mathcal{A}}), \cdots, (H^N_{\mathcal{A}}, I^N_{\mathcal{A}})\}$ where~$H^i_{\mathcal{A}},\, I^i_{\mathcal{A}}\subseteq Q$, $\forall i =1,2,\cdots, N$.
An infinite run~$q_0q_1q_2\cdots$ of~$\mathcal{A}$ is \emph{accepting} if there exists \emph{at least one} pair~$(H^i_{\mathcal{A}},\, I^i_{\mathcal{A}})\in \text{Acc}_{\mathcal{A}}$ such that~$\exists n\geq 0$, it holds~$\forall m\geq n,\, q_m\notin H^i_{\mathcal{A}}$ and~$\overset{\infty}{\exists} k\geq 0$,~$q_k\in I^i_{\mathcal{A}}$, where~$\overset{\infty}{\exists}$ stands for~``existing infinitely many''.
Namely, this run should intersect with~$H^i_{\mathcal{A}}$ \emph{finitely} many times while with~$I^i_{\mathcal{A}}$ \emph{infinitely} many times.
There are translation tools~\cite{klein2007ltl2dstar} to obtain~$\mathcal{A}_{\varphi}$ given~$\varphi$ with complexity~$2^{2^{\mathcal{O}(|\varphi|\log |\varphi|)}}$.

\section{Problem Formulation}\label{sec:problem-formulate}
In this section, we formally define the considered system model, the task specification,
the safe-return constraints, and the complete problem formulation.

\subsection{{Labeled MDP}}\label{subsec:probMDP}
{In order to model different workspace properties,
  the definition of a standard MDP~\cite{puterman2014markov} is extended
  with a labeling function over the states}, namely:
\begin{equation}\label{eq:mdp}
\mathcal{M} \triangleq \big(X, \, U,\, D,\, p_D, \, AP, \,L,\, c_D,\, x_0\big),
\end{equation}
where~$X$ is the finite state space;
$U$ is the finite control action space
(with a slight abuse of notation, $U(x)$ also denotes the set of control actions \emph{allowed} at state~$x\in X$);
$D \triangleq \{(x,\, u)\,|\, x\in X,\, u\in U(x)\}$ is the set of allowed transitions
as the possible state-action pairs;
$p_D\colon X\times U \times X \rightarrow {[0, 1]}$ is the transition probability function for each transition in $D$,
such that ~$\sum_{\check{x}\in X}p_D(x,u,\check{x}) = 1$, $\forall (x,\,u) \in D$;
$c_D \colon D \rightarrow \mathbb{R}^{>0}$ is the cost function associated with an action;
$AP$ is a set of atomic propositions as the properties of interest;
{$L \colon X \rightarrow 2^{AP}$ returns the properties held at each state (or simply labels);}
and lastly~$x_0\in X$,~$l_0\in L(x_0)$ are the initial states and labels.
Such models can be obtained by combining the robot motion controller with the environment model.

\begin{remark}\label{remark:prob-label}
  {
    The above model can be extended to probabilistically-labeled MDPs,
    as proposed in our previous work~\cite{guo2018probabilistic}.
    It can incorporate probabilistic labels at each state, which are useful
    for modeling uncertain environments.
    Moreover, for large-scale systems, such model~$\mathcal{M}$ can often
    be constructed algorithmically from data without much manual inputs,
    by combining the workspace data and the robot motion model,
    e.g., the office blueprint including walls and doors,
    and the satellite depth image of mountains and valleys.
    More details can be found in the simulation of Sec.~\ref{sec:case}.
  }
  \hfill $\blacksquare$
\end{remark}

\begin{example}\label{expample:mdp}
The robot motion for
the search-and-rescue mission is modeled as follows:
$X$ is a set of regions with the desired granularity~\cite{lavalle2006planning};
$U$ maps to the  navigation function among these regions;
$p_D$ is computed based on the feasibility of such navigation;
$L$ is the features such as different rooms and whether there are victims.\hfill $\blacksquare$
\end{example}

\subsection{Task Specification}
Different from the simple navigation task,
we take into account a more complex task specification
as LTL formulas $\varphi$ over the same atomic propositions~$AP$ from~\eqref{eq:mdp} above.
They can be used to specify both temporal and spatial requirements over the system,
of which the exact syntax and properties are given in Sec.~\ref{subsec:LTL}.
They are general enough to specify most high-level tasks such as surveillance, safety, service and response.
Many useful templates can be found in related work~\cite{hasanbeig2019reinforcement, laurenti2020formal, guo2015multi, vasile2017minimum, schillinger2019hierarchical, guo2016task, kantaros2020reactive}.

\begin{example}\label{example:task}
The task to surveil regions $r_1,r_2,r_3$ infinitely often can be specified as $\varphi=(\Box\Diamond r_1) \wedge (\Box\Diamond r_2)\wedge (\Box\Diamond r_3)$;
The task to pick an object from region $r_1$ and drop it at $r_2$ is given by
$\varphi=\Diamond ((\texttt{pick} \wedge r_1) \wedge \Diamond (\texttt{drop} \wedge r_2))$;
The task to provide supply at $r_1$ once a certain material is detected low is given by
$\varphi= \Box (\texttt{low} \rightarrow \Diamond(\texttt{supply} \wedge r_1))$.
\hfill $\blacksquare$
\end{example}

At stage~$T\geq 0$,
the robot's past path is given by~$X_T= x_0 x_1 \cdots x_T \in X^{(T+1)}$,
the past sequence of observed labels is given by~$L_T=l_0 l_{1}\cdots l_T\in (2^{AP})^{(T+1)}$
and the past sequence of control actions is~$U_T=u_0 u_1 \cdots u_T \in U^{(T+1)}$.
{It should hold that~$p_D(x_t, u_t, x_{t+1})>0$ and~$l_t=L(x_t)$, $\forall t\geq 0$.}
The complete past is then given by~$R_T=x_0l_0u_0 \cdots x_Tl_Tu_T$.
Denote by~$\boldsymbol{X}_T$, $\boldsymbol{L}_T$ and~$\mathbf{R}_T$ the set of all possible past sequences of states, labels, and runs up to stage~$T$.
{For brevity, $X_T \triangleq  R_T|_{X} $ and $L_T \triangleq R_T|_{L} $ denote the sequence of states and labels associated with the run~$R_T$.}
{For infinite sequences, $T=\infty$ and~$\boldsymbol{X}_\infty$, $\boldsymbol{L}_\infty$,~$\mathbf{R}_\infty$ denote the set of all infinite sequences of states, labels, and runs, respectively.}
{A \emph{finite-memory} policy is defined as~$\boldsymbol{\mu}=\mu_0\mu_1\cdots\mu_T$.}
The control policy at stage~$t\geq 0$ is given by~$\mu_t:\mathbf{R}_t \times U\rightarrow [0, \,1]$, $\forall t\geq 0$.
Denote by~$\overline{\boldsymbol{\mu}}$ the set of all such finite-memory policies.
Given a control policy~$\boldsymbol{\mu}\in \overline{\boldsymbol{\mu}}$,
the underlying system~$\mathcal{M}$ evolves as a Markov Chain (MC), denoted by~$\mathcal{M}|{\boldsymbol{\mu}}$.
{The probability measure on the smallest~$\sigma$-algebra,
over all possible infinite sequences within~$\mathcal{M}|{\boldsymbol{\mu}}$ that contain~$R_{T}$,
is the unique measure~\cite{baier2008principles}:
$Pr_{\mathcal{M}}^{\boldsymbol{\mu}}(R_T)= \prod_{t=0}^{T} \, p_D(x_t,\,u_t,\,x_{t+1}) \cdot {\mu_t(R_t,u_t)}$},
where~$\mu_t(R_t,u_t)$ is defined as the probability of choosing action~$u_t$ given the past run $\mathbf{R}_t$.
{Then, the probability of~$\mathcal{M}$ satisfying~$\varphi$ under a finite-memory policy~$\boldsymbol{\mu}$  is defined by}:
{
  \begin{equation}\label{eq:satisfy}
     \textbf{Sat}_{\mathcal{M}}^{\boldsymbol{\mu}} \triangleq {Pr}_{\mathcal{M}}^{\boldsymbol{\mu}}(\varphi) \triangleq
      {Pr}_{\mathcal{M}}^{\boldsymbol{\mu}}\big{(} R_{\infty}\in  \mathbf{R}_{\mathcal{M}}^{\boldsymbol{\mu}} \,|
     \, R_{\infty}|_{L} \models \varphi\big{)},
\end{equation}
}
{where $\mathbf{R}_{\mathcal{M}}^{\boldsymbol{\mu}}\subset \mathbf{R}_{\infty}$ is the set of all infinite runs of system~$\mathcal{M}$ under policy~${\boldsymbol{\mu}}$;
 $R_{\infty}|_{L}$ is the infinite sequence of labels associated with a run $R_{\infty}$ in~$\mathbf{R}_{\mathcal{M}}^{\boldsymbol{\mu}}$};
and the satisfaction relation~``$\models$'' is introduced in Sec.~\ref{subsec:LTL}.
Namely, the \emph{satisfiability} equals to the probability of all infinite runs whose associated labels satisfy the task.
More details on the probability measure can be found in~\cite{baier2008principles}.
Moreover, the \emph{cost} of policy $\boldsymbol{\mu}$ over~$\mathcal{M}$ is given by the expected mean cost of these infinite runs, namely:
\begin{equation}\label{eq:total-cost}
\begin{split}
\textbf{Cost}_{\mathcal{M}}^{\boldsymbol{\mu}}&\triangleq \mathbb{E}_{R_\infty\in \mathbf{R}_{\mathcal{M}}^{\boldsymbol{\mu}}}\{\textbf{Cost}(R_\infty)\}\\
&\triangleq \mathbb{E}_{R_\infty\in \mathbf{R}_{\mathcal{M}}^{\boldsymbol{\mu}}}\{\lim_{t\rightarrow \infty} \sum_{t=0}^{\infty} \frac{1}{t}\,c_D(x_t,u_t)\},
\end{split}
\end{equation}
where $\textbf{Cost}(R_\infty)$ is the \emph{mean} total cost~\cite{puterman2014markov} of an infinite run $R_{\infty}$;
and $c_D(\cdot)$ is the cost of applying $u_t$ and $x_t$ from~\eqref{eq:mdp}.
{Given only the task, our previous work in~\cite{guo2018probabilistic} proposes
  an approach to optimize the above cost by formulating two dependent Linear Programs for the plan
prefix and suffix.}

\begin{remark}\label{remark:cost}
Un-discounted cost summation is often used for the
stochastic shortest path problems~\cite{bertsekas1995neuro, polychronopoulos1996stochastic},
which are not suitable here as many general LTL tasks require liveness property
over  \emph{infinite} runs. \hfill $\blacksquare$
\end{remark}

\subsection{Safe-Return Constraints}\label{subsec:safety}
As discussed in Sec.~\ref{sec:introduction} and~\ref{sec:related-work},
traditionally safety is defined as the avoidance of unsafe states,
see~\cite{lavalle2006planning},
which are \emph{state-dependent} and mostly pre-defined offline.
Despite its intuitiveness,
it has serious drawbacks in scenarios where
safety depends on whether the system can follow a certain strategy to return to the safe states,
or where unsafe states can only be determined online during execution.
Such safety measure is now \emph{policy-dependent} and covers the traditional notion.

Formally, we consider the safe-return requirements~$\varphi_{\texttt{r}}$
specified as LTL formulas with the following format:
\begin{equation}\label{eq:safe-task}
  \varphi_{\texttt{r}}= \varphi^1_{\texttt{r}} \wedge \Diamond \Box\, \varphi^2_{\texttt{r}},
 \end{equation}
where~$\varphi^1_{\texttt{r}}$ are syntactically co-safe LTL (sc-LTL) formulas
over the same propositions~\cite{baier2008principles, pnueli1981temporal};
$\varphi^1_{\texttt{r}}$ denotes the transient requirement
{while $\varphi^2_{\texttt{r}}=\bigvee_{l\in L_s} l $, which contains the set of labels~$L_s\subset AP$ associated with a set of safe states that the system should \emph{stay} in the end}.
Note that sc-LTL formulas can be satisfied by a good prefix of finite length~\cite{pnueli1981temporal}.
Similar to~\eqref{eq:satisfy},
the probability of~$\mathcal{M}$ satisfying~$\varphi_{\texttt{r}}$ under a finite-memory policy $\boldsymbol{\mu}_{\texttt{r}}\in \overline{\boldsymbol{\mu}}$ is given by:
{\begin{equation}\label{eq:sat-return}
  \begin{split}
    \textbf{Sat}_{\mathcal{M}}^{\boldsymbol{\mu}_{\texttt{r}}}\triangleq Pr_{{\mathcal{M}}}^{\boldsymbol{\mu}_{\texttt{r}}}(\varphi_{\texttt{r}})
    \triangleq &{Pr}_{\mathcal{M}}^{\boldsymbol{\mu}_{\texttt{r}}}\big{(} R_\infty \in \mathbf{R}_{\mathcal{M}}^{\boldsymbol{\mu}_{\texttt{r}}} \,|\, \\
    &L_1 L^\omega_2 \models \varphi_{\texttt{r}},\, R_\infty|_L = L_1 L^\omega_2\big{)},
\end{split}
\end{equation}}
{where~$\mathbf{R}_{\mathcal{M}}^{\boldsymbol{\mu}_{\texttt{r}}}$ is the set of infinite runs of system~$\mathcal{M}$ under the policy~$\boldsymbol{\mu}_{\texttt{r}}$;
  one such run is denoted by~$R_\infty$ and its projection~$R_\infty|_L$ has the prefix-suffix format of~$L_1 L^\omega_2$,
  where~$L_1$ is a finite prefix that satisfies~$\varphi^1_{\texttt{r}}$ and
  $L^\omega_2$ is a cyclic suffix that satisfies~$\varphi^2_{\texttt{r}}$ as defined in~\eqref{eq:safe-task}}.

\begin{example}\label{example:safety}
  {For the search-and-rescue mission, it crucial that the robot can return to and stay at its base station via
  the designated exit, e.g., $\varphi_{\texttt{r}}= \Diamond \texttt{ex} \wedge \Diamond \Box \,\texttt{bs}$}.\hfill $\blacksquare$
\end{example}

Note that other safety constraints, such as collision avoidance, charging often, and emergency response, should be included in the task $\varphi$ rather than safe-return constraints~$\varphi_{\texttt{r}}$,
as in~\cite{hasanbeig2019reinforcement, laurenti2020formal, guo2015multi, schillinger2018simultaneous,
  tumova2013least, vasile2017minimum}.
More importantly, there are now \emph{two} finite-memory policies:
$\boldsymbol{\mu}_{\texttt{o}}$ called the \emph{outbound} policy that drives the robot to satisfy task $\varphi$;
and $\boldsymbol{\mu}_{\texttt{r}}$ called the \emph{return} policy that ensures the safety constraints $\varphi_{\texttt{r}}$.
{Due to the existence of the outbound policy, the satisfiability of system~$\mathcal{M}$ w.r.t. the safe-return requirements~$\varphi_{\texttt{r}}$ in~\eqref{eq:sat-return} is modified as follows:
{\begin{equation}\label{eq:sat-return-new}
  \begin{split}
  &\textbf{Safe}_{\mathcal{M}}^{\boldsymbol{\mu}_{\texttt{o}},\boldsymbol{\mu}_{\texttt{r}}} \triangleq
    \textbf{Sat}_{\mathcal{M}'_{\texttt{o}}}^{\boldsymbol{\mu}_{\texttt{r}}}
    = Pr_{\mathcal{M}'_{\texttt{o}}}^{\boldsymbol{\mu}_{\texttt{r}}}(\varphi_{\texttt{r}})\\
    &={Pr}_{\mathcal{M}'_{\texttt{o}}}^{\boldsymbol{\mu}_{\texttt{r}}}\big{(} R_\infty \in \mathbf{R}_{\mathcal{M}'_{\texttt{o}}}^{\boldsymbol{\mu}_{\texttt{r}}}\,|\, R_\infty|_L \models \varphi_{\texttt{r}}\big{)},
  \end{split}
\end{equation}
  where~$\mathcal{M}'_{\texttt{o}}$ is the modified model with an initial state distribution
  generated by system~$\mathcal{M}$ under the outbound policy~$\boldsymbol{\mu}_{\texttt{o}}$;
the word~$R_\infty|_L$ follows the prefix-suffix format as defined in~\eqref{eq:sat-return}.
Thus, the safe-return constraints are defined as follows}.

\begin{definition}\label{def:safety}
Given system $\mathcal{M}$, an outbound policy~$\boldsymbol{\mu}_{\texttt{o}}$ is called $\chi_{\texttt{r}}$-\emph{safe} if there exists a \emph{return} policy~$\boldsymbol{\mu}_{\texttt{r}}$ such that the probability of $\mathcal{M}$ satisfying $\varphi_{\texttt{r}}$ is lower-bounded:
\begin{equation}\label{eq:safe}
  \textbf{Safe}_{\mathcal{M}}^{\boldsymbol{\mu}_{\texttt{o}},\boldsymbol{\mu}_{\texttt{r}}} \geq \chi_{\texttt{r}},
\end{equation}
{or \emph{equivalently}: given $\boldsymbol{\mu}_{\texttt{o}}$ and~$\mathcal{M}$, $\exists \boldsymbol{\mu}_{\texttt{r}}$, s.t. $\textbf{Sat}_{\mathcal{M}|\boldsymbol{\mu}_{\texttt{o}}}^{\boldsymbol{\mu}_{\texttt{r}}} \geq \chi_{\texttt{r}}$, where $\chi_{\texttt{r}}\in [0,\,1]$ is the given safety bound.} \hfill $\blacksquare$
\end{definition}

It is worth clarifying that the task requirement~$\varphi$ and safe-return
constraints~$\varphi_\texttt{r}$ are \emph{fundamentally} different from the
multi-objective tasks considered in~\cite{forejt2011quantitative, etessami2007multi}.
This is due to the fact that in most cases there does not exist \emph{any}
policy~$\boldsymbol{\mu}$ that can satisfy~$\varphi$ and~$\varphi_\texttt{r}$
simultaneously, no matter how their relative priorities are set.
For instances, consider the surveillance task~$\varphi$ in Example~\ref{example:task}
and the safety constraint~$\varphi_\texttt{r}$ in Example~\ref{example:safety}.
They are mutually exclusive as  $\varphi$ requires it to surveil several regions
infinitely often,
while~$\varphi_\texttt{r}$ requires the system to return and stay at the base.

\subsection{Problem Statement}\label{subsec:problem}
\begin{problem}\label{prob:main}
Given the labeled MDP~$\mathcal{M}$ from~\eqref{eq:mdp},
the task~$\varphi$ and the safe-return requirement~$\varphi_{\texttt{r}}$,
our goal is to synthesize the outbound policy~$\boldsymbol{\mu}_{\texttt{o}}$ \emph{and} the return policy~$\boldsymbol{\mu}_{\texttt{r}}$
that solve the constrained optimization below:
\begin{equation}\label{eq:objective}
\begin{split}
&\min_{\boldsymbol{\mu}_{\texttt{o}},\boldsymbol{\mu}_{\texttt{r}}\in \overline{\boldsymbol{\mu}}} \;\; \textbf{Cost}_{\mathcal{M}}^{\boldsymbol{\mu}_{\texttt{o}}}\\
&\quad \text{s.t.} \quad \textbf{Sat}_{\mathcal{M}}^{\boldsymbol{\mu}_{\texttt{o}}} \geq \chi_{\texttt{o}}\;\; \text{and} \;\; \textbf{Safe}_{\mathcal{M}}^{\boldsymbol{\mu}_{\texttt{o}},\boldsymbol{\mu}_{\texttt{r}}} \geq \chi_{\texttt{r}},
\end{split}
\end{equation}
where $\chi_{\texttt{o}},\chi_{\texttt{r}}\in [0,\,1]$ are given lower bounds for satisfiability and safety in~\eqref{eq:satisfy} and~\eqref{eq:safe}, respectively;
the overall cost $\textbf{Cost}_{\mathcal{M}}^{\boldsymbol{\mu}_{\texttt{o}}}$ for the task is defined in~\eqref{eq:total-cost}.
\hfill $\blacksquare$
\end{problem}

{The above formulation has two implications: first, the synthesis of the outbound and inbound policies are coupled.
  The overall cost can not be optimized without ensuring \emph{both} the task satisfiability and the safe-return constraint;
  second, the system can evolve either under the outbound policy~$\boldsymbol{\mu}_{\texttt{o}}$
  or the return policy~$\boldsymbol{\mu}_{\texttt{r}}$, of which the switch is triggered
  by an \emph{external} ``return'' request.}
Furthermore, the lower bounds~$\chi_{\texttt{o}},\,\chi_{\texttt{r}}$ can {affect} the obtained policy greatly.
For a highly-uncertain workspace,~$\chi_{\texttt{o}}$ should be set low
while~$\chi_{\texttt{r}}$ should be high,
as no valid outbound {policies} exist if~$\chi_{\texttt{o}}$ is set too high.
On the other hand, for fairly certain models, both~$\chi_{\texttt{o}},\chi_{\texttt{r}}$
can be set high.
{More detailed discussions can be found in the numerical studies later in Sec.~\ref{sec:case}.}

\begin{remark}\label{remark:safety}
Most related work synthesizes only \emph{one} policy to satisfy $\varphi$ and $\varphi_{\texttt{r}}$ simultaneously, see~\cite{guo2018probabilistic,  hasanbeig2019reinforcement, laurenti2020formal, guo2015multi, schillinger2019hierarchical}.
This is however not directly possible as discussed after Def.~\ref{def:safety}, since the safe-constraints $\varphi_{\texttt{r}}$ are conflicting with $\varphi$ in most cases;
Second, the outgoing policy~$\boldsymbol{\mu}_{\texttt{o}}$ depends on the property of the return policy~$\boldsymbol{\mu}_{\texttt{r}}$ as in~\eqref{eq:safe},
thus can not be synthesized independently.
{Lastly, as elaborated earlier, only~$\boldsymbol{\mu}_{\texttt{o}}$ is {activated} during execution, while~$\boldsymbol{\mu}_{\texttt{r}}$ may
{never} be activated if not requested.}
In other words, $\varphi_\texttt{r}$ serves more as a constraint, instead of an actual task.
\hfill $\blacksquare$
\end{remark}

\begin{figure}[t]
\centering
   \includegraphics[width = 0.49\textwidth]{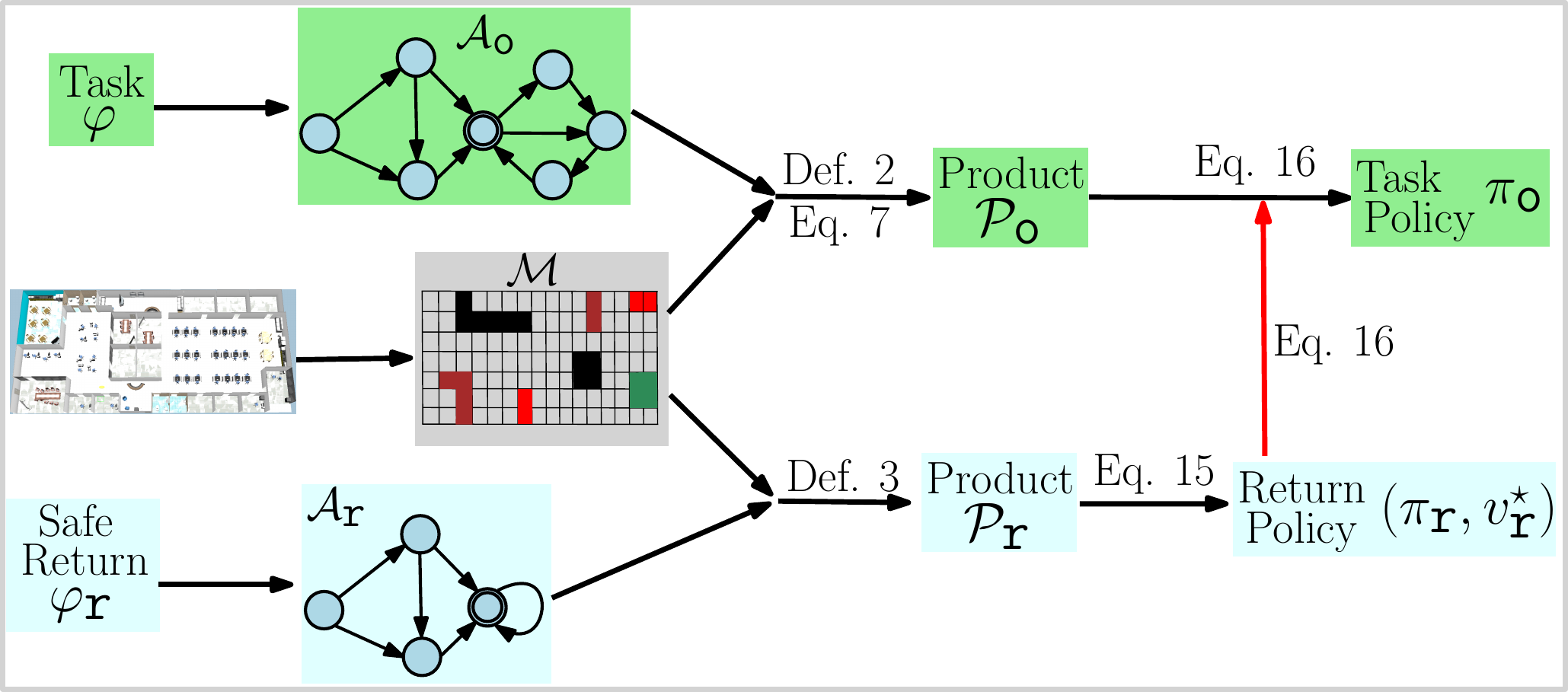}
  \caption{Framework of the proposed baseline solution, which includes constructing
the complete product automata~$\mathcal{P}_{\texttt{r}}$ and~$\mathcal{P}_{\texttt{o}}$,
and then solving two sequential optimizations.}
\label{fig:product_framework}
\end{figure}

\section{Theoretical Analyses and {Baseline Solutions}}\label{sec:analysis}
In this section,  we first provide the theoretical analyses on how the task and safe-return constraints in~\eqref{eq:objective}
can be re-formulated as constrained reachability problems in the respective product automata.
Based on these results,
we propose the baseline solution that solves two sequential optimizations using linear programming (LP) within these product automata,  as summarized in Fig.~\ref{fig:product_framework},
Lastly, we show that this solution quickly becomes intractable as the system size grows.

\subsection{Product Automaton and AMECs}\label{subsec:product}
As introduced in Sec.~\ref{subsec:dra}, we can construct the DRA~$\mathcal{A}_{\texttt{o}}$ associated with the task formula~$\varphi$ via the translation tools~\cite{klein2007ltl2dstar}.
Denote it by~$\mathcal{A}_{\texttt{o}}=(Q, \,2^{AP},\, \delta,\, q_{0},\,\text{Acc}_{\mathcal{A}})$,
where the detailed notations are omitted here.
Then we can construct a product automaton~\cite{baier2008principles} between~$\mathcal{M}$ and~$\mathcal{A}_{\texttt{o}}$.

\begin{definition}\label{def:product-o}
The product $\mathcal{P}_{\texttt{o}}\triangleq \mathcal{M}\times \mathcal{A}_{\texttt{o}}$ is a 7-tuple:
\begin{equation}\label{eq:prod-o}
\mathcal{P}_{\texttt{o}}=(S,\,U,\,E,\,p_E,\,c_E,\, s_0,\,\text{Acc}_{\mathcal{P}}),
\end{equation}
{where the state~$S\subseteq X\times Q$ satisfies~$\langle x,\,q \rangle \in S$, $\forall x \in X$ and~$\forall q\in Q$; the action set~$U$ is the same as in~\eqref{eq:mdp} and~$U(s)=U(x)$,~$\forall s=\langle x, q\rangle \in S$;} $E=\{(s,u)\,|\, s\in S,\, u\in U(s)\}$; {the transition probability~$p_E \colon S\times U \times S\rightarrow {[0,\,1]}$ is defined by}
\begin{equation}\label{eq:new_pro}
p_E\big{(}\langle x,q\rangle,\,u,\, \langle \check{x},\check{q}\rangle \big{)} =  p_D(x,\,u,\,\check{x}),
\end{equation}
{where (i) $\langle x, q\rangle,\, \langle \check{x}, \check{q}\rangle \in S$;
  (ii)~$(x,u)\in D$; and (iii)~$\check{q}= \delta(q,\,L(x))$}.
The label~$l$ fulfills the condition from~$q$ to~$\check{q}$ in~$\mathcal{A}_{\texttt{o}}$;
the cost function~$c_E \colon E\rightarrow \mathbb{R}^{>0}$ is given by~$c_E\big{(}\langle x,q\rangle,\,u \big{)}=c_D(x,u)$, $\forall {(}\langle x, q\rangle,\,u {)}\in E$;
{the initial state is~$s_0=\langle x_0, q_0 \rangle \in S$};
the accepting pairs are defined as~$\text{Acc}_{\mathcal{P}}=\{(H^i_{\mathcal{P}},\, I^i_{\mathcal{P}}), i=1,\cdots,N\}$, where~$H^i_{\mathcal{P}}=\{\langle x,q\rangle \in S\,|\, q\in H^i_{\mathcal{A}}\}$ and~$I^i_{\mathcal{P}}=\{\langle x,q\rangle \in S\,|\, q\in I^i_{\mathcal{A}}\}$,~$\forall i=1,\cdots,N$. \hfill $\blacksquare$
\end{definition}

The product~$\mathcal{P}_{\texttt{o}}$ computes the intersection between the traces of~$\mathcal{M}$ and the words of~$\mathcal{A}_{\texttt{o}}$,
to find the admissible robot behaviors that satisfy the task~$\varphi$.
Furthermore, the Rabin accepting condition of~$\mathcal{P}_{\texttt{o}}$ is the same as in Sec.~\ref{subsec:dra}. To transform this condition into equivalent graph properties,  we first compute the
accepting maximum end components (AMECs) of~$\mathcal{P}_{\texttt{o}}$ associated with its accepting pairs~$\text{Acc}_{\mathcal{P}}$.
Denote by~$\Xi_{acc}=\{S'_1,\, U'_1), \cdots (S'_{C},\, U'_{C})\}$ the set of AMECs associated with~$\text{Acc}_{\mathcal{P}}$,
where~$S'_{c}\subset S$ and~$U'_c:S'_c \rightarrow 2^{U}$,~$\forall c=1,\cdots,C$.
Simply speaking, $S'_c$ is the set of states the robot should converge to,
while $U'_c$ specifies the allowed actions at each state $s\in S'_c$
to remain inside $S'_c$.
Note that $S'_{c_1}\cap S'_{c_2}=\emptyset$, $\forall c_1,c_2=1,\cdots,C$.
Denote by
\begin{equation}\label{eq:AMECs}
S^{\textup{\texttt{o}}}_{\Xi}\triangleq \bigcup\limits_{c=1}^{C} S'_c,
\end{equation}
where $(S'_c,U'_c)\in \Xi_{acc}$ the union of all AMECs associated with~$\mathcal{P}_{\texttt{o}}$.
We omit the derivation of $\Xi_{acc}$ here
and refer the readers to Definitions 10.116, 10.117 and 10.124 of~\cite{baier2008principles} for theoretical details and~\cite{git_mdp_tg} for software implementation.

\begin{definition}\label{def:product-r}
Let $\mathcal{A}_{\texttt{r}}$ be the DRA associated with the safe-return constraints $\varphi_{\texttt{r}}$.
Then the product $\mathcal{P}_{\texttt{r}}\triangleq \mathcal{M}\times \mathcal{A}_{\texttt{r}}$ can be constructed analogously as $\mathcal{P}_{\texttt{o}}$ above,
of which the details are omitted here. \hfill $\blacksquare$
\end{definition}

Note that we omit the subscripts for all elements in $\mathcal{P}_{\texttt{o}}$ and $\mathcal{P}_{\texttt{r}}$ above for clarity.
Subscripts will be added whenever necessary to distinguish the same elements in different product automata.
Furthermore, the control policies for the product automata $\mathcal{P}_{\texttt{o}}$ and $\mathcal{P}_{\texttt{r}}$ are denoted by $\boldsymbol{\pi}_{\texttt{o}}$ and $\boldsymbol{\pi}_{\texttt{r}}$, respectively.
Note that due to the deterministic nature of DRA,
a policy in the product automaton, e.g., $\boldsymbol{\pi}_{\texttt{o}}$ and $\boldsymbol{\pi}_{\texttt{r}}$,
can be \emph{uniquely} mapped to a policy, e.g., $\boldsymbol{\mu}_{\texttt{o}}$ and $\boldsymbol{\mu}_{\texttt{r}}$ in $\mathcal{M}$, and vice versa.

\subsection{Task Satisfiability Reformulation}\label{subsec:task-reformulate}

Given the product~$\mathcal{P}_{\texttt{o}}$,
the following theorem is commonly used in related work~\cite{guo2018probabilistic, belta2017formal, ding2014optimal, ding2011mdp, forejt2011quantitative} to convert the task satisfiability to the reachability in~$\mathcal{P}_{\texttt{o}}$.

\begin{theorem} \label{theorem:task-bound}
The probability of $\varphi$ being satisfied by $\mathcal{M}$ under policy $\boldsymbol{\mu}_{\textup{\texttt{o}}}$ at stage $0$ can be re-formulated as:
\begin{equation}\label{eq:theo-sat-bound}
\textup{\textbf{Sat}}_{\mathcal{M}}^{\boldsymbol{\mu}_{\textup{\texttt{o}}}} = \mathbb{E}_{\widetilde{\mathcal{P}}_{\textup{\texttt{o}}}}^{\boldsymbol{\pi}_{\textup{\texttt{o}}}}\,
\left\{\sum_{t=0}^{\infty}\, b_{s_t,S^{\textup{\texttt{o}}}_{\Xi}}\right\},
\end{equation}
where~$\widetilde{\mathcal{P}}_{\textup{\texttt{o}}}\triangleq (1-b_{s,S^{\textup{\texttt{o}}}_{\Xi}})\cdot{\mathcal{P}}_{\textup{\texttt{o}}}$,
$b_{s,S^{\textup{\texttt{o}}}_{\Xi}}\triangleq \mathds{1}_{\{s \in S^{\textup{\texttt{o}}}_{\Xi}\}}$  is an indicator function;
$S^{\textup{\texttt{o}}}_{\Xi}$ is defined in~\eqref{eq:AMECs};
and $\boldsymbol{\pi}_{\textup{\texttt{o}}}$ is the policy for $\widetilde{\mathcal{P}}_{\textup{\texttt{o}}}$ corresponding to $\boldsymbol{\mu}_{\textup{\texttt{o}}}$.
\end{theorem}
\begin{proof}
It has been proven in~\cite{guo2018probabilistic, forejt2011quantitative,  baier2008principles}
that once the system $\mathcal{M}$ enters the union set of AMECs $S^{\textup{\texttt{o}}}_{\Xi}$ in $\mathcal{P}_{\textup{\texttt{o}}}$,
it can remain inside and satisfy the accepting condition of $\mathcal{P}_{\textup{\texttt{o}}}$ by following the transition conditions given by $U'_c$,
e.g., the Round-robin policy~\cite{baier2008principles} or the balanced policy~\cite{guo2018probabilistic}.
In other words,
the probability of satisfying $\varphi$ equals to the probability of entering $S^{\textup{\texttt{o}}}_{\Xi}$ \emph{at least once} in $\mathcal{P}_{\textup{\texttt{o}}}$.
Note that $\widetilde{\mathcal{P}}_{\textup{\texttt{o}}}$ has the same structure as~$\mathcal{P}_{\textup{\texttt{o}}}$.
But once the system reaches the set $S^{\textup{\texttt{o}}}_{\Xi}$ in~$\widetilde{\mathcal{P}}_{\textup{\texttt{o}}}$,
any further actions lead to a virtual ``end'' state with only self-loop.
Thus, the left-hand side of~\eqref{eq:theo-sat-bound} is computed by:
\begin{equation}\label{eq:sat}
\textbf{Sat}_{\mathcal{M}}^{\boldsymbol{\mu}_{{\textup{\texttt{o}}}}}= \mathbb{E}_{\mathcal{P}_{\textup{\texttt{o}}}}^{\boldsymbol{\pi}_{\texttt{o}}}
\,\Big\{\mathds{1}_{\{\exists t<\infty, s_t\in S^{\textup{\texttt{o}}}_{\Xi}\}}\Big\} = \mathbb{E}_{\widetilde{\mathcal{P}}_{\textup{\texttt{o}}}}^{\boldsymbol{\pi}_{\textup{\texttt{o}}}}  \,\left\{\sum_{t=0}^{\infty}b_{s_t,S^{\textup{\texttt{o}}}_{\Xi}}\right\},
\end{equation}
which completes the proof.
\end{proof}

\subsection{Safe-return Constraints Reformulation}\label{subsubsec:safety-reformulate}
The safe-return constraints in~\eqref{eq:safe} however have not been studied before in related work.
The following theorem is essential to re-formulate such constraints.

\begin{theorem} \label{theorem:safe-bound}
The safe-return constraints in~\eqref{eq:safe} for $\mathcal{M}$ under policies $\boldsymbol{\mu}_{\textup{\texttt{o}}}$ and $\boldsymbol{\mu}_{\textup{\texttt{r}}}$ at stage~$0$ can be re-formulated as:
\begin{subequations}\label{eq:theo-safe-bound}
\begin{align}
  &\textup{\textbf{Safe}}_{\mathcal{M}}^{\boldsymbol{\mu}_{\textup{\texttt{o}}},\boldsymbol{\mu}_{\textup{\texttt{r}}}} = \mathbb{E}_{\mathcal{P}_{\textup{\texttt{o}}}}^{\boldsymbol{\pi}_{\textup{\texttt{o}}}}\, \left\{\sum_{t=0}^{\infty}\, v_{\textup{\texttt{r}}}^\star(s^{\textup{\texttt{r}}}_t)\right\},\\
  &\text{where}\; v_{\textup{\texttt{r}}}^\star(s^{\textup{\texttt{r}}}_t) = \, \mathbb{E}_{\widetilde{\mathcal{P}}_{\textup{\texttt{r}}}}^{s^{\textup{\texttt{r}}}_{t},\boldsymbol{\pi}_{\textup{\texttt{r}}}}\left\{\sum_{\ell=t}^{\infty}  b_{s^{\textup{\texttt{r}}}_\ell,S^{\textup{\texttt{r}}}_{\Xi}}\right\},
\end{align}
\end{subequations}
where~$\widetilde{\mathcal{P}}_{\textup{\texttt{r}}}\triangleq (1-b_{s,S^{\textup{\texttt{r}}}_{\Xi}})\cdot{\mathcal{P}_{\textup{\texttt{r}}}}$,
$b_{s,S^{\textup{\texttt{r}}}_{\Xi}}\triangleq \mathds{1}_{\{s \in S^{\textup{\texttt{r}}}_{\Xi}\}}$ is an indicator function,
and~$S^{\textup{\texttt{r}}}_{\Xi}$ is defined in~\eqref{eq:AMECs};
$v_{\textup{\texttt{r}}}^\star \in [0,\, 1]$ is the value function of $\widetilde{\mathcal{P}}_{\textup{\texttt{r}}}$ under policy $\boldsymbol{\pi}_{\textup{\texttt{r}}}$;
the product state of $\mathcal{P}_{\textup{\texttt{o}}}$ at stage $t$ is denoted {by~$s_{t}=\langle x_t,q^{\textup{\texttt{o}}}_t\rangle$},
of which the associated product state in~$\widetilde{\mathcal{P}}_{\textup{\texttt{r}}}$ is given by {$s^{\textup{\texttt{r}}}_{t}=\langle x_t,q^{\textup{\texttt{r}}}_0\rangle$};
and $\boldsymbol{\pi}_{\textup{\texttt{o}}},\boldsymbol{\pi}_{\textup{\texttt{r}}}$ are the policies for~$\mathcal{P}_{\textup{\texttt{o}}}$ and $\mathcal{P}_{\textup{\texttt{r}}}$, corresponding to $\boldsymbol{\mu}_{\textup{\texttt{o}}},\boldsymbol{\mu}_{\textup{\texttt{r}}}$ for~$\mathcal{M}$, respectively.
\end{theorem}
\begin{proof}
The safe-return constraints can be expanded by \emph{splitting} the evolution of system~$\mathcal{M}$
before and after the time when the robot is requested to return,
then starts following the return policy.
Without loss of generality, denote by~$t\geq 0$ this time instant.
Thus,~$\mathcal{M}$ evolves under the outbound policy $\boldsymbol{\mu}_{\texttt{o}}$ to satisfy $\varphi$ between time $[0,\, t)$, then under the return policy $\boldsymbol{\mu}_{\texttt{r}}$ to satisfy $\varphi_{\texttt{r}}$ from time $\ell\in [t,\infty)$.
Thus, the definition of safety in~\eqref{eq:safe} can be expanded as follows:
\begin{equation}\label{eq:safe-bound-1}
\textbf{Safe}_{\mathcal{M}}^{\boldsymbol{\mu}_{\textup{\texttt{o}}},\boldsymbol{\mu}_{\textup{\texttt{r}}}} = \mathbb{E}_{\mathcal{P}_{\textup{\texttt{o}}}}^{\boldsymbol{\pi}_{\texttt{o}}} \,\left\{\sum_{t=0}^{\infty}\,  \textbf{Sat}_{\mathcal{M}}^{x_t, \boldsymbol{\mu}_{{\textup{\texttt{r}}}}}\right\},
\end{equation}
{where $s_{t}=\langle x_t,q^{\textup{\texttt{o}}}_t\rangle$} is the product state
 of $\mathcal{P}_{\textup{\texttt{o}}}$ at stage $t$;
and $\textbf{Sat}_{\mathcal{M}}^{x_t,\boldsymbol{\mu}_{{\textup{\texttt{r}}}}}$ is defined analogously to~\eqref{eq:theo-sat-bound} as the satisfiability of $\varphi_{\texttt{r}}$ under policy $\boldsymbol{\mu}_{\texttt{r}}$,
for system~$\mathcal{M}$ {but with a modified initial state $x_t$ at stage~$t$}.
Via the same argumentation as in Theorem~\ref{theorem:task-bound},
the probability of satisfying $\varphi_{\texttt{r}}$ equals to the probability of entering the union set $S^{\textup{\texttt{r}}}_{\Xi}$ of $\mathcal{P}_{\texttt{r}}$ at least once as time approaches infinity.
Then, consider that $\widetilde{\mathcal{P}}_{\textup{\texttt{r}}}$ defined above has the same structure as~$\mathcal{P}_{\textup{\texttt{r}}}$ except that once the system reaches $S^{\textup{\texttt{r}}}_{\Xi}$,
any further actions lead immediately to a virtual ``end'' state with only self-loop.
Thus it holds that:
\begin{equation}\label{eq:safe-bound-2}
  \begin{split}
\textbf{Sat}_{\mathcal{M}}^{x_t,\boldsymbol{\mu}_{{\textup{\texttt{r}}}}}&=\mathbb{E}_{\mathcal{P}_{\textup{\texttt{r}}}}^{s^{\textup{\texttt{r}}}_t,\boldsymbol{\pi}_{\texttt{r}}}\, \Big\{\mathds{1}_{\{\exists \ell\in [t,\,\infty),\, s^{\textup{\texttt{r}}}_\ell\in S^{\texttt{r}}_{\Xi}\}}\Big\}\\
  &=  \mathbb{E}_{\widetilde{\mathcal{P}}_{\textup{\texttt{r}}}}^{s^{\textup{\texttt{r}}}_t,\boldsymbol{\pi}_{\texttt{r}}}\,\left\{\sum_{\ell=t}^{\infty} b_{s^{\textup{\texttt{r}}}_\ell,S^{\texttt{r}}_{\Xi}}\right\} =v_{\textup{\texttt{r}}}^\star(s^{\textup{\texttt{r}}}_t),
  \end{split}
\end{equation}
{where $s^{\textup{\texttt{r}}}_t=\langle x_t,q^{\textup{\texttt{r}}}_0\rangle$} is the associated initial product state of $\mathcal{P}_{\textup{\texttt{r}}}$ at stage $t$.
This mapping is necessary as the states in $\mathcal{A}_{\textup{\texttt{o}}}$ and $\mathcal{A}_{\textup{\texttt{r}}}$ can be different;
and $v_{\textup{\texttt{r}}}^\star$ is by definition the value function of $\widetilde{\mathcal{P}}_{\textup{\texttt{r}}}$ under policy $\boldsymbol{\pi}_{\textup{\texttt{r}}}$.
Since $\boldsymbol{\mu}_{\textup{\texttt{r}}}$ optimizes the probability of satisfying $\varphi_{\texttt{r}}$,
it holds that its value function $v_{\textup{\texttt{r}}}^\star(s) \in [0, 1]$, $\forall s\in S_{\texttt{r}}$.
This completes the proof.
\end{proof}

Theorems~\ref{theorem:task-bound} and~\ref{theorem:safe-bound} provide theoretical guarantees on the re-formulation of both constraints in Problem~\ref{prob:main}.
More specifically,
Theorem~\ref{theorem:task-bound} states that
the task satisfiability can be ensured by limiting the reachability of $S^{\textup{\texttt{o}}}_{\Xi}$ within~$\widetilde{\mathcal{P}}_{\textup{\texttt{o}}}$ under policy $\boldsymbol{\pi}_{\texttt{o}}$.
Theorem~\ref{theorem:safe-bound} states that
the safe-return constraint can be evaluated as the expected \emph{rewards} of $\widetilde{\mathcal{P}}_{\texttt{o}}$ under policy~$\boldsymbol{\pi}_{\texttt{o}}$,
where the rewards are given by the value function of $\widetilde{\mathcal{P}}_{\texttt{r}}$ under the  return policy $\boldsymbol{\pi}_{\texttt{r}}$ by maximizing the reachability of $S^{\textup{\texttt{r}}}_{\Xi}$.
\setlength{\textfloatsep}{5pt}
\begin{algorithm}[t]
  \caption{Baseline Solution} \label{alg:direct}
  \LinesNumbered
  \KwIn{$\mathcal{M}$, $(\varphi, \chi_{\texttt{o}})$, $(\varphi_{\texttt{r}}, \chi_{\texttt{r}})$.}
  \KwOut{$(\mathcal{P}_{\texttt{r}}, \boldsymbol{\pi}_{\texttt{r}}, v_{\textup{\texttt{r}}}^\star)$, $(\mathcal{P}_{\texttt{o}}, \boldsymbol{\pi}_{\texttt{o}})$.}
  Build product $\mathcal{P}_{\texttt{r}}$ and its AMECs $S_{\Xi}^{\texttt{r}}$ \tcp*{Def.~\ref{def:product-r}}
  Synthesize policy~$\boldsymbol{\pi}_{\texttt{r}}$ and its value function $v_{\textup{\texttt{r}}}^\star$ \tcp*{Eq.~\eqref{eq:new-problem-safe}}
  Build product $\mathcal{P}_{\texttt{o}}$ and its AMECs $S_{\Xi}^{\texttt{o}}$ \tcp*{Def.~\ref{def:product-o}}
  Synthesize policy $\boldsymbol{\pi}_{\texttt{o}}$ given~$v_{\textup{\texttt{r}}}^\star$  \tcp*{Eq.~\eqref{eq:new-problem}}
\end{algorithm}

\subsection{Baseline Solution and Computational Complexity}\label{subsec:complexity}

Consequently,
the original problem after re-formulation can be solved by two sequential {optimizations}:
(i) synthesize the optimal return policy~$\boldsymbol{\pi}_{\texttt{r}}$ by solving the optimization:
\begin{equation}\label{eq:new-problem-safe}
    \max_{\boldsymbol{\mu}_{\texttt{r}}\in \overline{\boldsymbol{\mu}}} \;\; \textbf{Sat}_{\mathcal{M}}^{\boldsymbol{\mu}_{{\textup{\texttt{r}}}}} = \max_{\boldsymbol{\pi}_{\texttt{r}}\in \overline{\boldsymbol{\pi}}} \;\; \mathbb{E}_{\widetilde{\mathcal{P}}_{\textup{\texttt{r}}}}^{\boldsymbol{\pi}_{\texttt{r}}}\,\left\{\sum_{\ell=0}^{\infty} b_{s_\ell,S^{\texttt{r}}_{\Xi}}\right\},
\end{equation}
i.e., to {maximize} the reachability of $S^{\texttt{r}}_{\Xi}$ in~$\widetilde{\mathcal{P}}_{\textup{\texttt{r}}}$.
This amounts to solving a standard MDP problem without constraints, e.g., via LP.
Since $\widetilde{\mathcal{P}}_{\texttt{r}}$ becomes a Markov Chain under policy~$\boldsymbol{\pi}_{\texttt{r}}$,
the associated value function~$v_{\textup{\texttt{r}}}^\star$ can be easily computed, e.g., via value iteration;
(ii) synthesize the optimal outbound policy~$\boldsymbol{\pi}_{\texttt{o}}$ by solving the following constrained optimization:
\begin{equation}\label{eq:new-problem}
\begin{split}
&\min_{\boldsymbol{\pi}_{\texttt{o}}\in \overline{\boldsymbol{\pi}}} \;\; \textbf{Cost}_{\mathcal{P}_{\textup{\texttt{o}}}}^{\boldsymbol{\pi}_{\texttt{o}}}\\
\text{s.t.} &\quad \mathbb{E}_{\widetilde{\mathcal{P}}_{\textup{\texttt{o}}}}^{\boldsymbol{\pi}_{\textup{\texttt{o}}}}\,
  \left\{\sum_{t=0}^{\infty}\, b_{s_t,S^{\textup{\texttt{o}}}_{\Xi}}\right\}\geq \chi_{\texttt{o}},\\
  &\quad \mathbb{E}_{\widetilde{\mathcal{P}}_{\textup{\texttt{o}}}}^{\boldsymbol{\pi}_{\textup{\texttt{o}}}}\, \left\{\sum_{t=0}^{\infty}\, v_{\textup{\texttt{r}}}^\star(s^{\textup{\texttt{r}}}_t)\right\} \geq \chi_{\texttt{r}},
\end{split}
\end{equation}
where the relevant notations are defined in~\eqref{eq:theo-sat-bound}  and~\eqref{eq:theo-safe-bound}.
The above problem amounts to a \emph{constrained} MDP problem~\cite{altman1999constrained, bertsekas1991analysis}.
As a result, a methods similar to our earlier work~\cite{guo2018probabilistic} can be applied to synthesize the optimal prefix and suffix policies of $\boldsymbol{\pi}_{\texttt{o}}$.
Briefly speaking, a constrained LP is formulated over the occupancy measure over all state-action pairs,
such that (i) the summation of occupancy measure entering $S^{\textup{\texttt{o}}}_{\Xi}$ is larger than $\chi_{\texttt{o}}$;
(ii) the summation of occupancy measure multiplied by the associated value function over all states is larger than $\chi_{\texttt{r}}$;
and (iii) the summation of occupancy measure multiplied by the transition cost within $S^{\textup{\texttt{o}}}_{\Xi}$ is minimized.
The resulting outbound policy $\boldsymbol{\pi}_{\texttt{o}}$ is a stochastic policy over $\mathcal{P}_{\textup{\texttt{o}}}$,
while the return policy $\boldsymbol{\pi}_{\texttt{r}}$ is a  stochastic policy over $\widetilde{\mathcal{P}}_{\textup{\texttt{r}}}$.
We refer the readers to the supplementary material for the detailed LP formulation and policy derivation.
{It is worth noting that the mean cost minimization only applies to the
  outgoing policy~$\boldsymbol{\mu}_{\texttt{o}}$ not the return policy~$\boldsymbol{\mu}_{\texttt{o}}$
  as formulated in~\eqref{eq:objective}.}
The above procedure is summarized in Fig.~\ref{fig:product_framework} and Alg.~\ref{alg:direct}.

\begin{figure}
\begin{center}
  \begin{minipage}[t]{0.7\linewidth}
\strut\vspace*{-\baselineskip}\newline
    \centering
   \includegraphics[width = 0.9\textwidth]{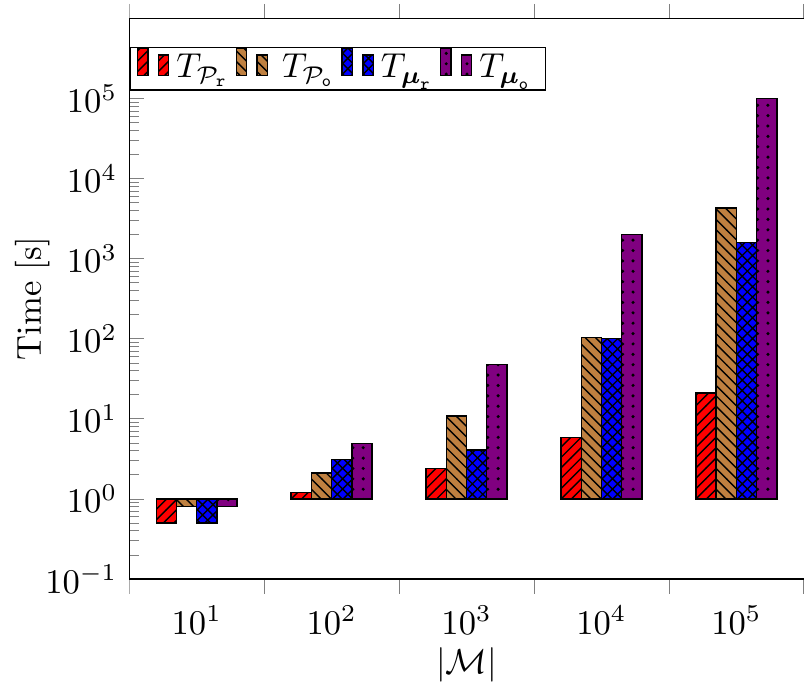}
  \end{minipage}%

  \begin{minipage}[t]{0.9\linewidth}
\strut\vspace*{-\baselineskip}\newline
\vspace{0.5cm}
\resizebox{\textwidth}{!}{
  \begin{tabular}{ccccc} \hline
    $|\mathcal{M}|$ & $|\mathcal{P}_{\texttt{r}}|$ & $|\mathcal{P}_{\texttt{o}}|$ & $T_{\boldsymbol{\mu}_{\texttt{r}}} [s]$ & $T_{\boldsymbol{\mu}_{\texttt{o}}} [\si{\second}]$\\ \hline
\addlinespace[0.3em]
  $(1\texttt{e}2,8\texttt{e}2)$ & $(5\texttt{e}2,4\texttt{e}3)$ & $(4\texttt{e}3,5\texttt{e}4)$ & $0.1$ & $0.9$ \\
\addlinespace[0.3em]
  $(5\texttt{e}2,4\texttt{e}3)$ & $(3\texttt{e}3,3\texttt{e}4)$ & $(2\texttt{e}4,2\texttt{e}5)$ & $2.7$ & $5.9$ \\
\addlinespace[0.3em]
  $(2\texttt{e}3,2\texttt{e}4)$ & $(1\texttt{e}4,1\texttt{e}5)$ & $(6\texttt{e}4,6\texttt{e}5)$ & $4.1$ & $48$ \\
\addlinespace[0.3em]
  $(7\texttt{e}3,7\texttt{e}4)$ & $(5\texttt{e}4,4\texttt{e}5)$ & $(2\texttt{e}5,3\texttt{e}6)$ & $1\texttt{e}2$ & $2\texttt{e}3$ \\
\addlinespace[0.3em]
  $(3\texttt{e}4,3\texttt{e}5)$ & $(2\texttt{e}5,3\texttt{e}6)$ & $(9\texttt{e}5,8\texttt{e}6)$ & $2\texttt{e}3$ & N/A \\
\addlinespace[0.3em]
  $(1\texttt{e}5,2\texttt{e}6)$ & $(7\texttt{e}5,8\texttt{e}6)$ & N/A & N/A & N/A\\
\hline
  \end{tabular}
}
\end{minipage}
\end{center}
\vspace{-0.2in}
\caption{\textbf{Top}: Size of product $\mathcal{P}_{\texttt{r}}$, $\mathcal{P}_{\texttt{o}}$
(measured by the number of nodes and edges);
and computation time of polices $\boldsymbol{\mu}_{\texttt{r}}$, $\boldsymbol{\mu}_{\texttt{o}}$,
where~$a\texttt{e}b\triangleq a\times 10^b$.
Note that $10^4\si{\second}\approx 2.7\si{\hour}$,
whereas``N/A'' indicates either insufficient memory error or computation longer than $24$ hours;
\textbf{Bottom}: Plot of model computation time and solution time w.r.t. the size of the underlying $\mathcal{M}$.
}
\label{fig:small-complexity-compare}
\end{figure}

Note that the product~$\mathcal{P}_{\texttt{r}}$ has the approximate size of~$|2^{\varphi_{\texttt{r}}}|\cdot |\mathcal{M}|$,
of which the AMECs are computed in polynomial time.
The optimal return policy~$\boldsymbol{\pi}_{\texttt{r}}$ can be synthesized in polynomial time also due to LP.
Similar analysis holds also for $\mathcal{P}_{\texttt{r}}$.
Nonetheless, for high dimensional states, large workspaces and complex tasks considered in practice,
the LPs above can become \emph{intractable} to even construct, let alone to solve.
Some examples are given below to illustrate the computation time blow-up with increasing system size,
e.g., for a moderate size of $\mathcal{M}$ (around $10^5$ edges), it takes more than $24$ hours to synthesize the outgoing policy.

\begin{example}\label{example:complexity}
Consider the surveillance task in Example~\ref{example:task}, the associated DRA has $32$ states, $282$ edges and $1$ accepting pairs,
while the DRA associated with the safe-return task in Example~\ref{example:safety} has $7$ states, $14$ edges and $1$ accepting pairs.
Fig.~\ref{fig:small-complexity-compare} summarizes
how $\mathcal{P}_{\texttt{r}}$ and $\mathcal{P}_{\texttt{o}}$ increase in size,
when the underlying model~$\mathcal{M}$ grows.
It is apparent that the baseline solution quickly becomes intractable
as even formulating the LPs takes hours.
For MDPs with more than $10^6$ edges, neither the return policy nor the outbound policy can be computed within reasonable amount of memory and time.
\hfill $\blacksquare$
\end{example}

{
\subsection{{Alternative Baseline}}\label{subsec:alternative}
{Another straightforward but approximate method as an alternative baseline to the above baseline
is to build an extended model~$\widehat{\mathcal{M}}$
  of the original~$\mathcal{M}$ in~\eqref{eq:mdp} by adding another dimension},
  i.e.,~$\widehat{X}=\{\langle x,\, i\rangle, \forall x \in X, \, i \in \{0,\, 1\}\}$,
  where $i$ indicates whether a return has been requested.
More specifically, when~$i=0$, the system~$\widehat{\mathcal{M}}$ evolves according to the outbound
policy~$\boldsymbol{\mu}_{\texttt{o}}$;
when~$i=1$, the system~$\widehat{\mathcal{M}}$ evolves under the return policy~$\boldsymbol{\mu}_{\texttt{r}}$.
In this way, both the outbound and return policies might be integrated into \emph{one} policy as they act on different
states in~$\widehat{\mathcal{M}}$.
However, since this return request is an external signal and not controlled by the robot,
an \emph{approximation} would be to modify the transition probability of~$\mathcal{M}$ as follows:
(i) $\widehat{p}_D(\langle x,0\rangle, u, \langle \hat{x},\hat{i}\rangle)=p_D(x,u,\hat{x})\cdot 0.5$,
$\forall \hat{i}\in \{0,1\}$ and $\forall (x,u)\in D$;
(ii) $\widehat{p}_D(\langle x,1\rangle, u, \langle \hat{x},1\rangle)=p_D(x,u,\hat{x})$,
$\forall (x,u)\in D$.
In other words, the system can transit non-deterministically within level~$i=0$ or from level~$i=0$ to~$i=1$ at each time step,
but not in the reversing order.
Then, via the multi-objective probabilistic model checking algorithm proposed in~\cite{Forejt12Pareto}
and available in PRISM~\cite{kwiatkowska2011prism},
one common policy, denoted by~$\widehat{\boldsymbol{\mu}}$, can be synthesized
to satisfy both the objectives
that{~$Pr_{\widehat{\mathcal{M}}}^{\widehat{\boldsymbol{\mu}}}(\varphi_{\texttt{r}})\geq \chi_{\texttt{r}}$ for~$i=0$
and~$Pr_{\widehat{\mathcal{M}}}^{\widehat{\boldsymbol{\mu}}}(\varphi)\geq \chi_{\texttt{o}}$ for~$i=1$,}
while minimizing the rewards in~\eqref{eq:objective}.
More details can be found in the supplementary material.}

{Since the extended model~$\widehat{\mathcal{M}}$ has~$2|X|$ nodes and~$3|U|$ edges,
  the associated product automaton, denoted by~$\widehat{\mathcal{P}}$,
  has roughly the size of~$3|U|\cdot 2^{|\varphi|} \cdot 2^{|\varphi_{\texttt{r}}|}$,
  see~\cite{Forejt12Pareto} for the method to construct~$\widehat{\mathcal{P}}$.
  The computational complexity of this alternative solution can be estimated
  as~$\mathcal{O}((3|U|\cdot 2^{|\varphi|} \cdot 2^{|\varphi_{\texttt{r}}|})^{N_c})$,
i.e., at least $(3^{N_c}\cdot 2^{|\varphi_{\texttt{r}}|})$ times the size of the constrained optimization in~\eqref{eq:new-problem},
with~$N_c>2.3$ being the currently-known best algorithm to solve a LP.
In other words, this alternative solution is at least a magnitude more expensive than
the baseline solution proposed in the section.
This difference can be even more significant if the specifications~$\varphi_{\texttt{r}}$, $\varphi$ are complex,
see Sec.~\ref{subsubsec:exp-compare} for detailed comparisons.}

\section{Proposed Hierarchical Solution}\label{sec:solution}
To tackle the intractable complexity of the baseline solution,
the main hierarchical solution is proposed in this section.
As illustrated in Fig.~\ref{fig:proposed_framework},
a two-step paradigm similar to the baseline solution is followed.
{It follows the framework of semi-MDPs and options as proposed in~\cite{sutton1999between}.
The major difference is that two feature-based symbolic and temporal abstraction of system~$\mathcal{M}$ as semi-MDPs are constructed for the task specification and the safe-return constraints, respectively.
Given these models, a hierarchical planning algorithm is proposed to synthesize both the return policy and outgoing policy as options that solve the original problem}.

\subsection{Hierarchical Planning for Safe-return Constraints}\label{subsec:safe-return-plan}

As discussed before,
the safe-return policy has to be synthesized first along with its value function,
which is then used to compute the outbound policy.
Thus, we first describe how the safe-return policy can be synthesized {hierarchically}.

\begin{figure}[t]
\centering
   \includegraphics[width = 0.49\textwidth]{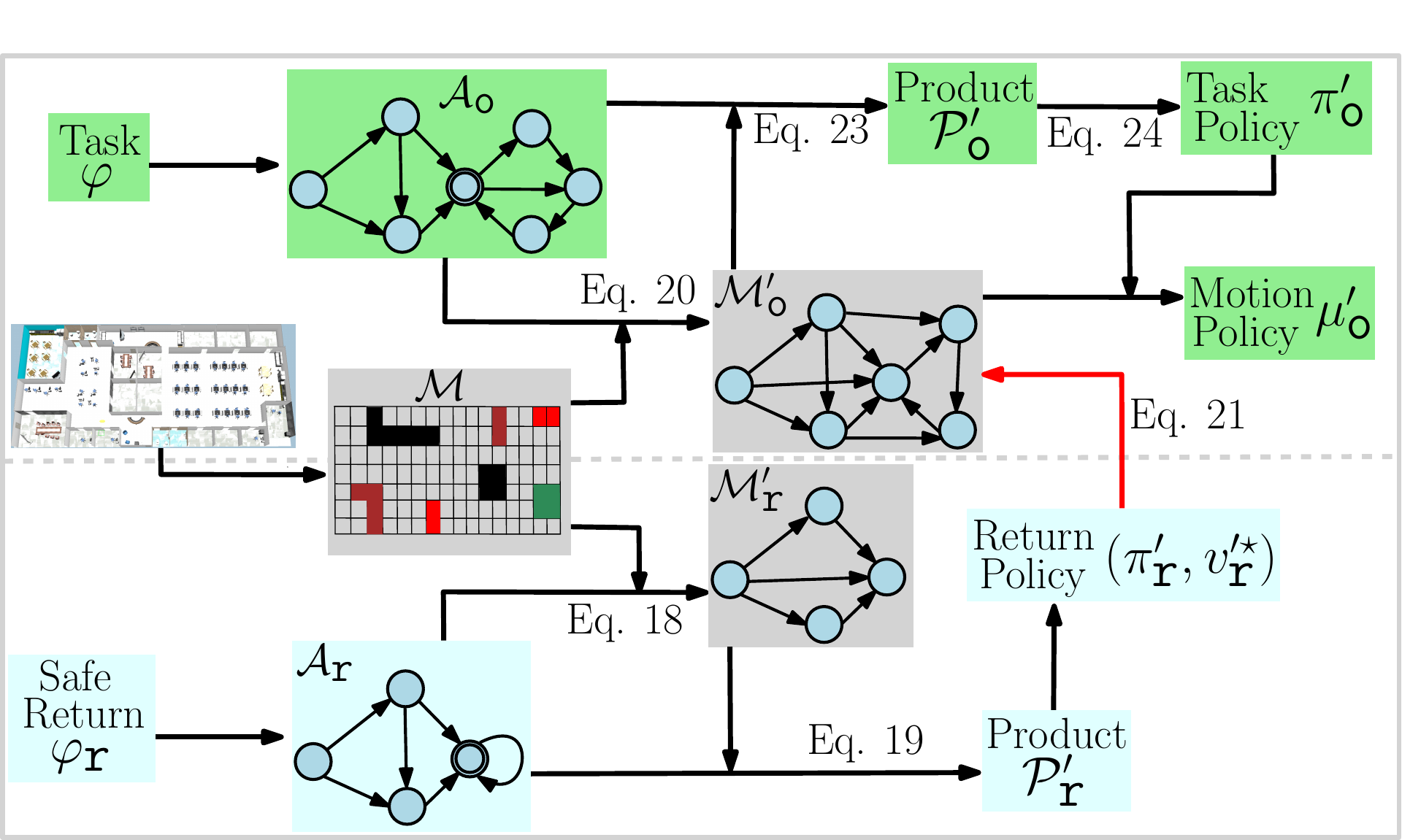}
  \caption{Illustration of the proposed hierarchical planning framework.
Compared with the baseline solution in Fig.~\ref{fig:product_framework},
two feature-based abstraction models~$\mathcal{M}'_{\texttt{o}}$, $\mathcal{M}'_{\texttt{r}}$
are constructed first for the task specification and the safe-return constraints, respectively.}
\label{fig:proposed_framework}
\end{figure}

\subsubsection{{Labeled Semi-MDPs for Safe-return Constraints}}\label{subsubsec:abstract-safe}
Given the DRA $\mathcal{A}_{\texttt{r}}=(Q, \,2^{AP},\, \delta,\, q_{0},\,\text{Acc}_{\mathcal{A}})$
associated with~$\varphi_{\texttt{r}}$,
we can compute the set of \emph{effective} features:
\begin{equation}\label{eq:effectve-safe}
  \Theta_{\texttt{r}} = \{\theta \in 2^{AP}\,|\, \exists q,\,\check{q}\in Q,\,
  \text{s.t.}\; (q,\,\theta,\check{q})\in \delta,\, q\neq \check{q}\},
\end{equation}
which includes any label inside~$\Theta_{\texttt{r}}$ that can drive a transition within $\mathcal{A}_{\texttt{r}}$, excluding self-transitions.
{Then, the feature-based abstraction of the original MDP~$\mathcal{M}$ for $\varphi_{\texttt{r}}$ as semi-MDPs,
  denoted by~$\mathcal{M}'_{\textup{\texttt{r}}}$, can be constructed as}
another labeled MDP, namely:
\begin{equation}\label{eq:abstract-mdp-safe}
\mathcal{M}'_{\textup{\texttt{r}}} \triangleq \big(X'_{\textup{\texttt{r}}}, \, U'_{\textup{\texttt{r}}},\, p'_{\textup{\texttt{r}},D}, \, AP_{\textup{\texttt{r}}}, \,L'_{\textup{\texttt{r}}}\big),
\end{equation}
where $X'_{\textup{\texttt{r}}}=\{x\in X\,|\, L(x) \cap \Theta_{\textup{\texttt{r}}} \neq \emptyset\}\subset X$ contains \emph{only} the states that can potentially lead to a transition within $\mathcal{A}_{\texttt{r}}$;
{$AP_{\textup{\texttt{r}}},\,L'_{\textup{\texttt{r}}}$ are defined analogously as in~\eqref{eq:mdp}};
$U'_{\textup{\texttt{r}}}$ is the set of {macro} actions that represent {symbolically} the underlying motion policies for each transition in $X'_{\textup{\texttt{r}}}$;
$p'_{\textup{\texttt{r}},D}: X'_{\textup{\texttt{r}}} \times U'_{\textup{\texttt{r}}} \times X'_{\textup{\texttt{r}}} \rightarrow [0,\, 1]$ is the transition probability for each transition.
The derivation of $p'_{\textup{\texttt{r}},D}$ is explained below.


\begin{problem}\label{problem:determine-path}
Determine $U'_{\textup{\texttt{r}}}$, $p'_{\textup{\texttt{r}},D}$ in~\eqref{eq:abstract-mdp-safe},
given~$\mathcal{M}$. \hfill $\blacksquare$
\end{problem}

For \emph{each} pair of states $(x_f,\,x_t)\in X'_{\textup{\texttt{r}}}\times X'_{\textup{\texttt{r}}}$,
the associated macro action is symbolically denoted by $(x_f,\,x_t)\in U'_{\textup{\texttt{r}}}$.
As illustrated in Fig.~\ref{fig:feature-abstraction}, the transition probability $p'_{\textup{\texttt{r}},D}$ is computed in three steps:
(i) Construct modified MDP $\mathcal{M}'_{\textup{\texttt{r}}}(x_f,\,x_t)$ from~$\mathcal{M}$,
such that state $x_f$ is the ``source'' state,
and all the other states in $X'_{\textup{\texttt{r}}}$ (including $x_t$) are the ``sink'' states.
Once the system enters any sink state it will stay there via self-loop with probability one and cost zero;
(ii) Find the motion policy in~$\mathcal{M}'_{\textup{\texttt{r}}}(x_f,\,x_t)$ that drives the system from $x_f$ to $x_t$ with the \emph{maximum} probability.
{Denote by $\boldsymbol{\mu}'_{\texttt{r}}(x_f,\,x_t)$ this specific motion policy, which is analogous to the \emph{options} in~\cite{sutton1999between}.}
Similar to the maximum reachability problem discussed in~\eqref{eq:new-problem-safe} before,
this can be readily solved by formulating the LP over the occupancy measures of each state-action pair in~$\mathcal{M}'_{\textup{\texttt{r}}}(x_f,\,x_t)$;
(iii) Given the policy~$\boldsymbol{\mu}'_{\texttt{r}}(x_f,\,x_t)$,
the underlying MDP $\mathcal{M}'_{\textup{\texttt{r}}}(x_f,\,x_t)$ becomes a Markov Chain (MC),
of which of asymptotic behavior is fully determined.
Thus, the transition probability $p'_{\textup{\texttt{r}},D}\left(x_f,\boldsymbol{\mu}'_{\texttt{r}}(x_f,\,x_t),x_\ell\right)$ from the source state $x_f$ to \emph{any other} sink state $x_\ell\in X'_{\textup{\texttt{r}}}$ is given by the final distribution over the sink states.
Note that if the LP above does not have a solution for the pair $(x_f,\,x_t)$,
it means $x_t$ {cannot} be reached from $x_f$.
Consequently, the transition probability $p'_{\textup{\texttt{r}},D}\left(x_f,\boldsymbol{\mu}'_{\texttt{r}}(x_f,\,x_t),x_\ell\right)=0$ for all $x_\ell \in X'_{\textup{\texttt{r}}}$.
The detailed formulation of the LP and computation of~$p'_{\textup{\texttt{r}},D}$ can be found in the supplementary material.

\begin{figure}[t]
\centering
   \includegraphics[width = 0.48\textwidth]{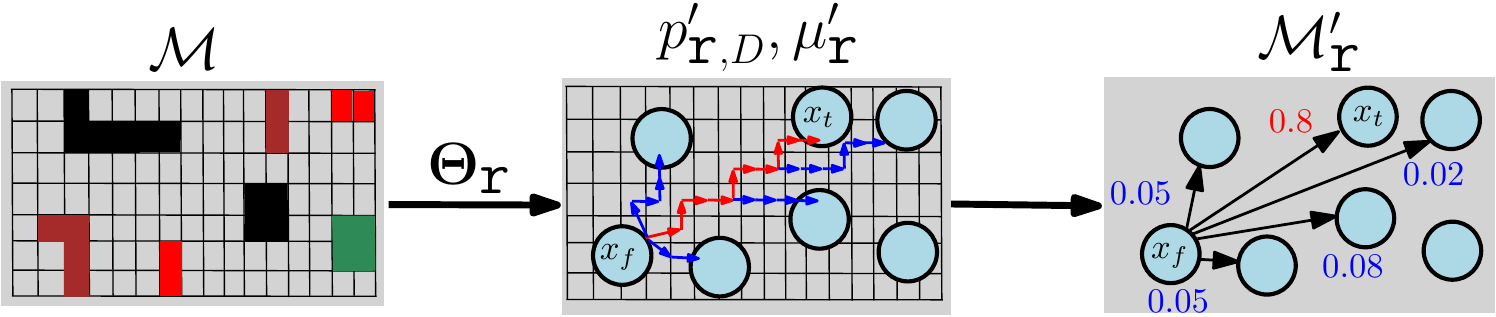}
  \caption{Illustration of feature-based abstraction as described in Sec.~\ref{subsubsec:abstract-safe}.
The associated motion policy~$\boldsymbol{\mu}'_{\texttt{r}}$ and the transition probability~$p'_{\texttt{r}, D}$ should be computed for each pair of transition $(x_f,x_t)$ in $\mathcal{M}'_{\texttt{r}}$.
Note that the ending states under the macro action $(x_f, x_t)$  are non-deterministic,
as shown in the middle and right figures.}
\label{fig:feature-abstraction}
\end{figure}

{Via the above three steps, the labeled semi-MDP~$\mathcal{M}'_{\texttt{r}}$ in~\eqref{eq:abstract-mdp-safe} can be constructed fully.}
Assume that the DRA~$\mathcal{A}_{\texttt{r}}$ has $N$ unique labels,
and within $\mathcal{M}$ there are at most $M$ regions of interest associated with each label,
then $\mathcal{M}'_{\texttt{r}}$ has maximum $M\cdot N$ nodes and less than $M^2\cdot N^2$ edges.
For large systems with few number of interested regions, this can be significantly less than the size of $\mathcal{M}$.

\subsubsection{Hierarchical Safe-Return Policy Synthesis}\label{subsubsec:safe-return-policy}
{Given the labeled semi-MDP~$\mathcal{M}'_{\texttt{r}}$ from~\eqref{eq:abstract-mdp-safe}}
and the DRA $\mathcal{A}_{\texttt{r}}$,
their product can be computed in the same way as in Def.~\ref{def:product-o}:
\begin{equation}\label{eq:abstract-safe-product}
\mathcal{P}'_{\texttt{r}}\triangleq \mathcal{M}'_{\texttt{r}} \times \mathcal{A}_{\texttt{r}}=(S'_{\texttt{r}},\,U'_{\texttt{r}},\,E'_{\texttt{r}},\,p'_{\texttt{r},E},\, s'_{\texttt{r},0},\,\text{Acc}'_{\texttt{r}}),
\end{equation}
of which the detailed notations are similarly defined as in~\eqref{eq:prod-o} and omitted here.
Note that the transition probability~$p'_{\texttt{r},E}$ is computed based on $p'_{\textup{\texttt{r}},D}$.
Given $\mathcal{P}'_{\texttt{r}}$ above, the optimal safe-return policy $\boldsymbol{\pi}'_{\texttt{r}}$ and the associated value function $v'^{\star}_{\texttt{r}}$ can be computed in the same way as computing $\boldsymbol{\pi}_{\texttt{r}}$ from $\mathcal{P}_{\texttt{r}}$,
which are described in Sec.~\ref{subsec:complexity}.
Furthermore, since~$\varphi_{\texttt{r}}$ is given as sc-LTL formulas,
the safe-return constraints are satisfied once the union set of AMECs~$S'_{\texttt{r},\Xi}$ above is reached.

\subsection{Hierarchical Planning for Tasks}\label{subsec:task-plan}
Similar to the return policy above,
the outbound policy can also be synthesized in a hierarchical way.
{Namely, a feature-based abstraction model as semi-MDPs for the task specification is firstly constructed},
which is safety-ensured as it directly incorporates the safe-return constraints.
Second, a hierarchical outbound policy is synthesized given the
product automaton between this abstraction model and the task automaton.

\subsubsection{{Safety-ensured Semi-MDPs for Tasks}}\label{subsubsec:abstract-task}

Different from~$\mathcal{M}'_{\texttt{r}}$ in~\eqref{eq:abstract-mdp-safe},
{the abstraction model for tasks as semi-MDPs should incorporate the safe-return constraints}.
First, the set of effective features within~$\mathcal{A}_{\texttt{o}}$ is
computed similarly as~\eqref{eq:effectve-safe},
denoted by~$\Theta_{\texttt{o}}$.
Then, the associated abstraction model as semi-MDPs is given by:
\begin{equation}\label{eq:abstract-mdp-task}
\mathcal{M}'_{\textup{\texttt{o}}} \triangleq \big(X'_{\textup{\texttt{o}}}, \, U'_{\textup{\texttt{o}}},\, p'_{\textup{\texttt{o}},D}, \, AP, \,L'_{\textup{\texttt{o}}},\,c'_{\textup{\texttt{o}},D}\big),
\end{equation}
{where the notations are defined analogously as in~\eqref{eq:abstract-mdp-safe}}.
However, the transition probability~$p'_{\textup{\texttt{o}},D}$ and cost~$c'_{\textup{\texttt{o}},D}$ are
computed differently to incorporate the safe-return constraints.

\begin{problem}\label{problem:determine-prob}
Determine $p'_{\textup{\texttt{o}},D}$, $c'_{\textup{\texttt{o}},D}$ of $\mathcal{M}'_{\texttt{o}}$,
given system~$\mathcal{M}$ and the value function~$v_{\texttt{r}}'^\star$
of~$\mathcal{P}'_{\texttt{r}}$. \hfill $\blacksquare$
\end{problem}

For \emph{each} pair of states $(x_f,\,x_t)\in X'_{\textup{\texttt{o}}}\times X'_{\textup{\texttt{o}}}$,
the associated macro action is symbolically denoted by $(x_f,\,x_t)\in U'_{\textup{\texttt{o}}}$.
{Then, its transition probability $p'_{\textup{\texttt{o}},D}$ is computed in  a similar procedure as~$p'_{\texttt{r},D}$ in Sec.~\ref{subsec:safe-return-plan}, namely:
(i) Construct the modified MDP~$\mathcal{M}'_{\texttt{o}}(x_f,\,x_t)$ from~$\mathcal{M}$ similar to~$\mathcal{M}'_{\texttt{r}}(x_f,\,x_t)$;}
(ii) {Find the motion policy~$\boldsymbol{\mu}'_{\texttt{o}}(x_f,\,x_t)$ as options that drives the system~$\mathcal{M}'_{\texttt{o}}(x_f,\,x_t)$ from~$x_f$ to $x_t$ with the \emph{maximum} probability},
while satisfying the safe-return constraint.
Theorem~\ref{theorem:safe-bound} proves that the safe-return constraint can be re-formulated as
the accumulated rewards w.r.t. the value function $v'^\star_{\texttt{r}}$.
Thus, the LP for computing $\boldsymbol{\mu}'_{\texttt{o}}(x_f,\,x_t)$ is revised
by adding another constraint:
\begin{equation}\label{eq:abstract-task-lp}
\mathbb{E}_{\mathcal{M}'_{\texttt{o}}(x_f,\,x_t)}^{\boldsymbol{\mu}'_{\textup{\texttt{o}}}}\, \left\{\sum_{t=0}^{\infty}\, v'^\star_{\texttt{r}}(s^{\texttt{r}}_t)\right\} \geq \chi_{\texttt{r}},
\end{equation}
{where~$s_t^{\texttt{r}}=\langle x_t, q^{\texttt{r}}_0\rangle$ is the product state
of~$\mathcal{P}'_{\texttt{r}}$ derived from~\eqref{eq:abstract-safe-product}};
$v'^\star_{\texttt{r}}(s^{\texttt{r}}_t)$ is the associated value function from~\eqref{eq:abstract-safe-product};
(iii) The transition probability $p'_{\textup{\texttt{o}},D}\left(x_f,\boldsymbol{\mu}'_{\texttt{o}}(x_f,\,x_t),x_\ell\right)$
is the convergent distribution of~$\mathcal{M}'_{\texttt{o}}(x_f,\,x_t)$
under~$\boldsymbol{\mu}'_{\texttt{o}}(x_f,\,x_t)$.
Furthermore, the cost function $c'_{\textup{\texttt{o}},D}$ is determined by:
\begin{equation}\label{eq:task-edge-cost}
c'_{\textup{\texttt{o}},D}(x_f,\,x_t) = \mathbb{E}_{\mathcal{M}'_{\texttt{o}}(x_f,\,x_t)}^{\boldsymbol{\mu}'_{\textup{\texttt{o}}}} \left\{\sum_{t=0}^{\infty}\, c_D(x_t,u_t)\right\},
\end{equation}
as the expected cost of reaching~$x_t$ from $x_f$ while following the motion policy $\boldsymbol{\mu}'_{\texttt{o}}(x_f,\,x_t)$.
Note if the constrained LP for solving $\boldsymbol{\mu}'_{\texttt{o}}(x_f,\,x_t)$ \emph{does not} have a solution,
it means that $x_t$ can not be reached from $x_f$ while satisfying the safe-return constraints in~\eqref{eq:abstract-task-lp}.
Then, the macro transition~$(x_f, x_t)$ is removed from $U'_{\texttt{o}}$ with no associated transition.
The detailed formulation of the LP and derivation of $p'_{\texttt{o},D},\, c'_{\texttt{o},D}$ can be found in the supplementary material.

\begin{remark}\label{remark:safe-policy}
{The motion policy~$\boldsymbol{\mu}'_{\texttt{o}}(x_f,\,x_t)$ above is an option that prioritizes the maximization of the probability of reaching~$x_f$ from~$x_t$},
instead of minimizing the expected cost $c'_{\textup{\texttt{o}},D}(\cdot)$ in~\eqref{eq:task-edge-cost}.
This could result in a loss of optimality regarding the original objective in~\eqref{eq:objective}.
{First of all}, as also mentioned in our earlier work~\cite{guo2018probabilistic},
there are often trade-offs when co-optimizing both terms.
Second, {the overall task satisfiability in~\eqref{eq:objective} is not incorporated in this \emph{local} semi-MDP},
rather only in the product automaton in the sequel.
Nonetheless, as shown later in the experiment, such loss is negligible, especially in contrast to the significant reduction in computation time.
\hfill $\blacksquare$
\end{remark}

\begin{lemma}\label{lemma:safe}
{The semi-MDP~$\mathcal{M}'_{\texttt{o}}$ is safety-ensured}.
Namely, for each transition $(x_f, x_t) \in U'_{\texttt{o}}$,
its associated motion policy $\boldsymbol{\mu}'_{\textup{\texttt{o}}}(x_f, x_t)$ and the return policy $\boldsymbol{\mu}'_{\textup{\texttt{r}}}$ satisfy the safe-return constraints by~\eqref{eq:safe}.
\end{lemma}
\begin{proof}
For each transition $(x_f, x_t) \in U'_{\texttt{o}}$,
the safe-return constraints are enforced directly by~\eqref{eq:abstract-task-lp} when synthesizing the motion policy $\boldsymbol{\mu}'_{\textup{\texttt{o}}}(x_f, x_t)$.
As shown in Theorem~\ref{theorem:safe-bound}, the expected accumulative value function of all states encountered during the execution of $\boldsymbol{\mu}'_{\textup{\texttt{o}}}(x_f, x_t)$ is equivalent to the safe-return probability in~\eqref{eq:safe}.
Thus, during the execution of any transition within~$\mathcal{M}'_{\texttt{o}}$, once requested, the robot can always return to the safe states with a probability larger than~$\chi_\texttt{r}$, by following the safe-return policy~$\boldsymbol{\mu}'_{\textup{\texttt{r}}}$.
\end{proof}

\subsubsection{Hierarchical Outbound Policy Synthesis}\label{subsubsec:task-policy}
{Given the safety-ensured semi-MDP~$\mathcal{M}'_{\texttt{o}}$},
its product with the DRA~$\mathcal{A}_{\texttt{o}}$ can be computed via following Def.~\ref{def:product-r}:
\begin{equation}\label{eq:abstract-task-product}
  \mathcal{P}'_{\texttt{o}}\triangleq \mathcal{M}'_{\texttt{o}} \times \mathcal{A}_{\varphi_\texttt{o}},
\end{equation}
of which the detailed notations are similarly defined as in~\eqref{eq:prod-o} and omitted here.
Note that its cost function follows from~\eqref{eq:task-edge-cost}.
Now, the objective is to find the optimal outbound policy over $\mathcal{P}'_{\texttt{o}}$ for the original planning problem.

\setlength{\textfloatsep}{5pt}
\begin{algorithm}[t]
  \caption{Hierarchical Planning Algorithm} \label{alg:all}
  \LinesNumbered
  \KwIn{$\mathcal{M}$, $(\varphi, \chi_{\texttt{o}})$, $(\varphi_{\texttt{r}}, \chi_{\texttt{r}})$.}
  \KwOut{$\left((\mathcal{M}'_{\texttt{r}},\boldsymbol{\mu}'_{\texttt{r}}), (\mathcal{P}'_{\texttt{r}}, \boldsymbol{\pi}'_{\texttt{r}}, v'^\star_{\texttt{r}})\right)$, $\left((\mathcal{M}'_{\texttt{o}},\boldsymbol{\mu}'_{\texttt{o}}), (\mathcal{P}'_{\texttt{o}}, (\boldsymbol{\pi}'_{\texttt{o},\texttt{pre}}, \boldsymbol{\pi}'_{\texttt{o},\texttt{suf}}))\right)$}
  \tcc{Safe-return policy, Sec.~\ref{subsec:safe-return-plan}}
Construct DRA~$\mathcal{A}_{\texttt{r}}$\;
  Build abstraction~$\mathcal{M}'_{\texttt{r}}$ and  policy $\boldsymbol{\mu}'_{\texttt{r}}$\;
  Compute product $\mathcal{P}'_{\texttt{r}}$ in~\eqref{eq:abstract-safe-product} \;
  Synthesize return policy~$\boldsymbol{\pi}'_{\texttt{r}}$, and value function $v'^\star_{\texttt{r}}$\;
  \tcc{Task policy, Sec.~\ref{subsec:task-plan}}
Construct DRA~$\mathcal{A}_{\texttt{o}}$\;
  Build abstraction~$\mathcal{M}'_{\texttt{o}}$ and policy~$\boldsymbol{\mu}'_{\texttt{o}}$, given $v'^\star_{\texttt{r}}$\;
  Compute product $\mathcal{P}'_{\texttt{o}}$ in~\eqref{eq:abstract-task-product}\;
  Synthesize outbound policy~$(\boldsymbol{\pi}'_{\texttt{o},\texttt{pre}}, \boldsymbol{\pi}'_{\texttt{o},\texttt{suf}})$ \;
\end{algorithm}

\begin{problem}\label{problem:max-task}
Determine the policy $\boldsymbol{\pi}'_{\texttt{o}}$ for $\mathcal{P}'_{\texttt{o}}$ such that the long-term cost in~\eqref{eq:total-cost} is minimized and the task satisfiability in~\eqref{eq:satisfy} is lowered bounded.
\hfill $\blacksquare$
\end{problem}

Notice that different from~$\mathcal{P}_{\texttt{o}}$,
the product~$\mathcal{P}'_{\texttt{o}}$ above \emph{inherits} the safety property from the abstraction model~$\mathcal{M}'_{\texttt{o}}$ as proved in Lemma~\ref{lemma:safe}.
Consequently, the safe-return constraints are not imposed as an additional requirement when synthesizing the outbound policy~$\boldsymbol{\pi}'_{\texttt{o}}$.
Since task $\varphi$ is given as general LTL formulas,
$\boldsymbol{\pi}'_{\texttt{o}}$ now consists of two parts:
the prefix~$\boldsymbol{\pi}'_{\texttt{o}, \texttt{pre}}$ that is executed once,
and the suffix~$\boldsymbol{\pi}'_{\texttt{o}, \texttt{suf}}$ that is executed infinite times.
Without the additional safe-return constraints,
the framework proposed in our earlier work~\cite{guo2018probabilistic} can be used
directly to synthesize~$\boldsymbol{\pi}'_{\texttt{o}}$.
For brevity, Alg.~1 in~\cite{guo2018probabilistic} is encapsulated as the following constrained optimization:
\begin{subequations}\label{eq:abstract-prod-task-lp}
\begin{align}
&\underset{\{\boldsymbol{\pi}'_{\texttt{o}, \texttt{pre}}, \boldsymbol{\pi}'_{\texttt{o}, \texttt{suf}}\}}{\boldsymbol{\min}} \; \textbf{Cost}_{\mathcal{P}'_{\texttt{o}}}^{\boldsymbol{\pi}'_{\texttt{o}, \texttt{suf}}} \label{eq:abstract-prod-task-lp-objective}\\
&\quad \textrm{s.t.}\quad \textbf{Sat}_{\mathcal{P}'_{\texttt{o}}}^{\boldsymbol{\pi}'_{\texttt{o}, \texttt{pre}}} \geq \chi_{\texttt{o}}, \label{eq:abstract-prod-task-lp-condition}
\end{align}
\end{subequations}
where the prefix policy $\boldsymbol{\pi}'_{\texttt{o}, \texttt{pre}}$ ensures that
the union of AMECs~$S'_{\texttt{o},\Xi}$ is reached with a probability larger than $\chi_{\texttt{o}}$,
while the suffix policy $\boldsymbol{\pi}'_{\texttt{o}, \texttt{suf}}$ ensures that
the system stays inside~$S'_{\texttt{o},\Xi}$ and the discounted cost in~\eqref{eq:objective} is minimized.
In the end, both policies are optimized together to find the best pair of initial state and accepting state.
The detailed LP formulation and policy derivation are given in the supplementary material.


\subsection{Algorithmic Summary and Online Execution} \label{subsec:summary}

The complete hierarchical planning algorithm is summarized in Alg.~\ref{alg:all} and illustrated in Fig.~\ref{fig:proposed_framework}.
Namely, it consists of two main parts:
{Lines~$1-4$ to construct the semi-MDP~$\mathcal{M}'_{\texttt{r}}$ and product~$\mathcal{P}'_{\texttt{r}}$},
and learn the safe-return policy~$\boldsymbol{\pi}'_{\texttt{r}}$ and the associated value function $v'^\star_{\texttt{r}}$;
{Lines~$5-8$ to construct the semi-MDP~$\mathcal{M}'_{\texttt{o}}$ and product~$\mathcal{P}'_{\texttt{o}}$},
and learn the outbound policy~$\boldsymbol{\pi}'_{\texttt{o}}$.
As shown in Fig.~\ref{fig:execution}, the outbound policy is executed hierarchically as discussed in the sequel.
The safe-return policy is activated only if the robot is requested to return.

During online execution, the outbound policy is executed
hierarchically as follows:
Starting from the initial state~$s'_{\texttt{o},k}$ and $k=0$, the prefix policy~$\boldsymbol{\pi}'_{\texttt{o},\texttt{pre}}$ is activated to determine the next action $(x_k, x^\star_{k+1})\in U'_{\texttt{o}}$.
Once $x^\star_{k+1}$ is given, {the option~$\boldsymbol{\mu}'_{\texttt{o},\texttt{pre}}(x_k, x^\star_{k+1})$ is activated to determine the next motion $(x_{k},x_{k,\ell})\in U$}, where $x_{k,\ell}\in X$ but $x_{k,\ell} \notin X'_{\texttt{o}}$.
The step~$\ell$ is increased until one of the sink states in $X'_{\texttt{o}}$ is reached,
which is denoted by $x_{k+1}$ and not necessarily $x^\star_{k+1}$ due to non-determinism.
Given the actual next state~$x_{k+1}$, the outbound policy~$\boldsymbol{\pi}'_{\texttt{o},\texttt{pre}}$ is activated again to determine the next action $(x_{k+1}, x^\star_{k+2})\in U'_{\texttt{o}}$.
This process repeats itself until the set~$S'_{\texttt{o},\Xi}$ of $\mathcal{P}'_{\texttt{o}}$ is reached.
Afterwards, the outbound policy switches to the suffix policy~$\boldsymbol{\pi}'_{\texttt{o},\texttt{suf}}$,
while the return policy~$\boldsymbol{\pi}'_{\texttt{r}}$ remains the same.
Last but not least, whenever the robot is requested to return,
the safe-return policy is executed hierarchically in an analogous way as above,
which however terminates after the system reaches the set~$S'_{\texttt{r},\Xi}$.


\begin{figure}[t]
\centering
   \includegraphics[width = 0.48\textwidth]{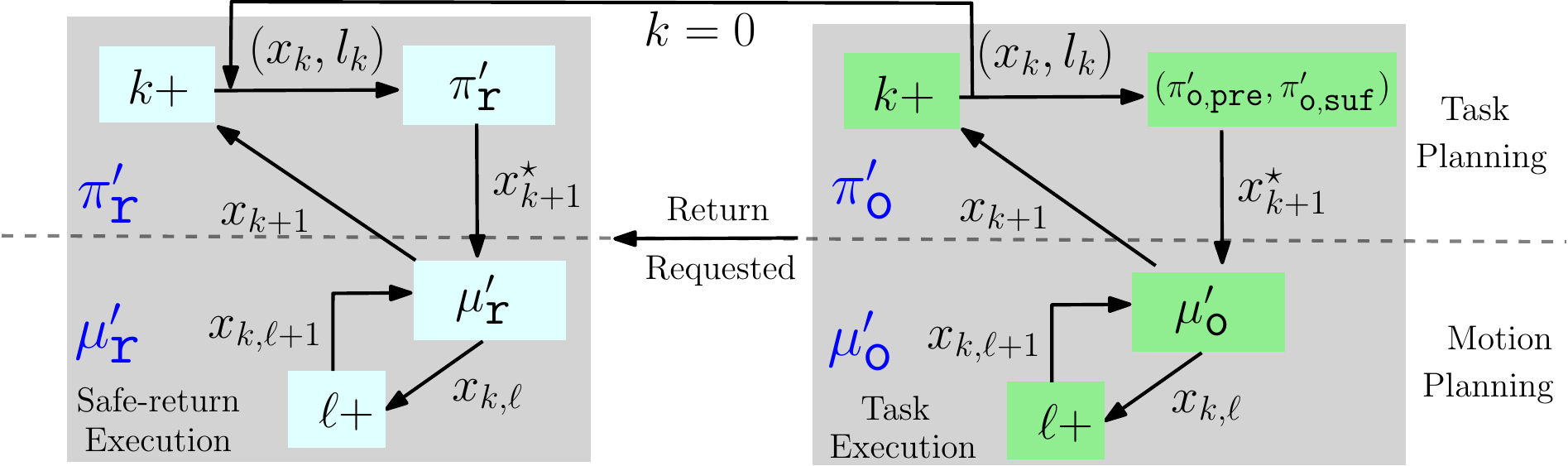}
  \caption{Illustration of the online execution of the hierarchical outbound policy~$(\boldsymbol{\pi}'_{\texttt{o}}, \boldsymbol{\mu}'_{\texttt{o}})$,
and the hierarchical safe-return policy $(\boldsymbol{\pi}'_{\texttt{r}}, \boldsymbol{\mu}'_{\texttt{r}})$ if requested.}
\label{fig:execution}
\end{figure}

\begin{theorem}\label{lem:safety-guarantee}
The safe-return policy $\boldsymbol{\pi}'_{\textup{\texttt{r}}}$  and the outbound policy~$\boldsymbol{\pi}'_{\textup{\texttt{o}}}$ fulfill the constraints in~\eqref{eq:objective} as imposed by Problem~\ref{prob:main},
and minimizes the long-term cost in~\eqref{eq:abstract-prod-task-lp-objective}.
\end{theorem}
\begin{proof}
First,
{it is proven in Lemma~\ref{lemma:safe} that the semi-MDP~$\mathcal{M}'_{\texttt{o}}$ is safety-ensured}.
In other words, during the execution of any transition within~$\mathcal{M}'_{\texttt{o}}$,
once requested the robot can always return to the safe states with a probability larger than~$\chi_\texttt{r}$,
{by following the safe-return option~$\boldsymbol{\mu}'_{\textup{\texttt{r}}}$}.
Thus, the outbound policy $\boldsymbol{\pi}'_{\textup{\texttt{o}}}$ derived by solving~\eqref{eq:abstract-prod-task-lp} for $\mathcal{P}'_{\texttt{o}}$ always satisfies the safe-return constraints.
Second, as proven in Theorem~6 of our earlier work~\cite{guo2018probabilistic},
the optimization in~\eqref{eq:abstract-prod-task-lp} guarantees the task satisfiability by co-optimizing the prefix and suffix of the outbound policy~$\boldsymbol{\pi}'_{\textup{\texttt{o}}}$.
In particular, it shows that the prefix policy~$\boldsymbol{\pi}'_{\texttt{o}, \texttt{pre}}$ ensures that
the union of AMECs~$S'_{\texttt{o},\Xi}$ is reached with a probability larger than $\chi_{\texttt{o}}$,
while the suffix policy $\boldsymbol{\pi}'_{\texttt{o}, \texttt{suf}}$ ensures that
the system stays inside~$S'_{\texttt{o},\Xi}$ and the long-term discounted cost is minimized.
Lastly, as mentioned in Remark~\ref{remark:safe-policy},
the transition cost $c'_{\textup{\texttt{o}},D}$ within $\mathcal{M}'_{\texttt{o}}$ might \emph{not} be equivalent to the minimum cost for each transition.
As a result, the discounted cost in~\eqref{eq:abstract-prod-task-lp-objective} is an approximation of the original cost in~\eqref{eq:objective}.
The exact gap between them is not quantified in this work and remains part of our ongoing research.
\end{proof}

\begin{remark}\label{remark:compare-old-new}
  {It is worth pointing out the algorithmic differences between the baseline method in Alg.~\ref{alg:direct} and
    the proposed hierarchical approach in Alg.~\ref{alg:all}.
    As shown in Fig.~\ref{fig:product_framework} and Fig.~\ref{fig:proposed_framework}, the baseline method directly synthesizes
    first the return policy~$\pi_{\texttt{r}}$ and then the outbound policy~$\pi_{\texttt{o}}$ over the original system~$\mathcal{M}$,
    while the proposed approach constructs first the symbolic and temporal abstractions~$\mathcal{M}'_{\texttt{r}}$, $\mathcal{M}'_{\texttt{o}}$ as semi-MDPs and their associated polices as options.
    Theses semi-MDPs are then used as the respective system model for synthesizing the high-level task policy and return policy.
    This hierarchical approach allows planning at different spatial and temporal levels over different system models,
    i.e., instead of a common model as in the baseline method.
  }
\hfill $\blacksquare$
\end{remark}

\section{Simulation and Experiment Study}\label{sec:case}
In this section, we present two case studies to validate the proposed framework:
the search-and-rescue mission in office environment after a disaster;
and the planetary exploration mission similar to the DLR SpaceBot Camp~\cite{droeschel2017continuous}.
All algorithms are implemented in Python3 and the main LP solver is
the Google Linear Optimization Solver ``\emph{Glop}'', see~\cite{glop}.
All computations are carried out on a laptop (3.06GHz Duo CPU and 12GB of RAM).
Detailed robot model and workspace descriptions can be found in the supplementary material.

\subsection{\textbf{Study One}: Search-and-Rescue Mission}\label{subsec:search-and-rescue}
For the first numerical study, we consider the deployment of a UGV into an office environment for a search-and-rescue mission after a disaster.

\subsubsection{Workspace and Task Description}\label{subsubsec:search-and-rescue-ws}
As shown in Fig.~\ref{fig:office-traj}, a search-and-rescue UGV of size $1\si{\meter}\times 0.5\si{\meter}$ is deployed to explore one floor of an office building, approximately of size $170\si{\meter}\times 80\si{\meter}$.
The system model~$\mathcal{M}$ is estimated by the floor plan and the robot mobility.
{More specifically, the states are given by a full discretization of the floor plan
  in 2D (with different granularity as described later);
  transition probabilities and costs are estimated by a simulated robot navigation model,
  e.g., to the adjacent regions via 4-way movements with orientation constraints.}
Note that the robot can drift side-ways while moving
or overshoot while rotating, yielding a probabilistic model;
and the states are labeled by the corresponding rooms,
while features such as victims or hazards are estimated manually.
{For instance},  the probability of having label human ($\texttt{hm}$) in offices is $0.9$,
whereas $0.1$ in storage rooms ($\texttt{st}$).
We refer the {readers} to the supplementary material for detailed construction of~$\mathcal{M}$.
{It is worth noting that given the workspace blueprint (including walls, doors and stairs)
  and the robot motion model, the model~$\mathcal{M}$ can be constructed algorithmically without manual inputs.}
Moreover, there are two potential risks during the mission:
first, the robot may be trapped in certain rooms due to one-way doors or some exits being blocked by debris;
second, the robot may descend any stairs but not ascend too steep stairs.

\begin{figure}[t]
  \begin{center}
\begin{minipage}[t]{0.9\linewidth}
    \includegraphics[width =1.02\textwidth]{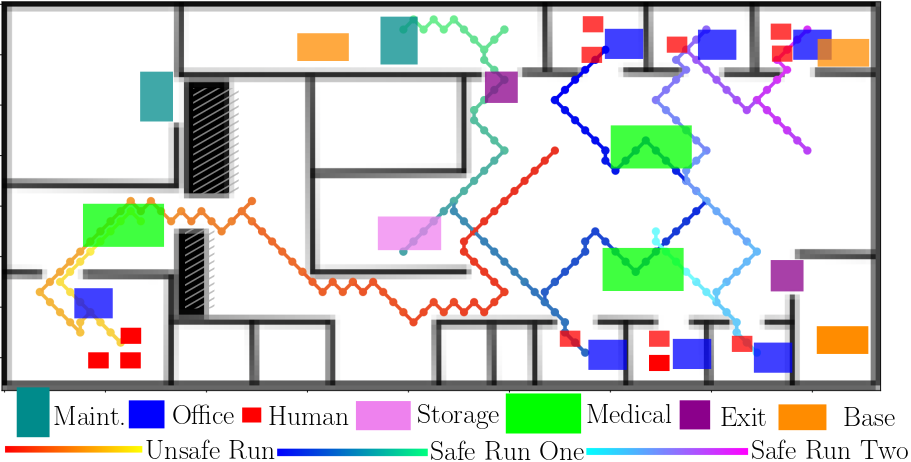}
\end{minipage}
\end{center}
  \caption{
Workspace model and examples of different runs under the outbound policy~$\boldsymbol{\mu}'_{\texttt{o}}$ and an unsafe policy (in red) without the safe constrains.}
\label{fig:office-traj}
\end{figure}

More specifically, {the search-and-rescue tasks include}:
(i) turn off switches at the maintenance room ($\texttt{mt}$);
(ii) visit storage rooms ($\texttt{st}$) to check for fire hazards and gas leakage;
(iii) search office rooms ($\texttt{of}$) for injured victims ($\texttt{hm}$) and bring them to the closest medical station ($\texttt{md}$).
The locations of these labels are shown in Fig.~\ref{fig:office-traj}.
They are specified as the following LTL formulas:
\begin{equation}\label{eq:serach-and-rescue-task}
  \begin{split}
    & \varphi=\; \Diamond \texttt{st} \wedge (\Diamond \bigvee\nolimits_{\ell\in L} \texttt{mt}_\ell) \\
    & \wedge (\Box \Diamond (\bigvee\nolimits_{i\in I} \texttt{of}_i \wedge \texttt{hm})) \wedge (\Box (\texttt{hm} \rightarrow (\neg \texttt{hm})\textsf{U} \texttt{md})),
    \end{split}
\end{equation}
where $I$ and $L$ are the sets of offices and maintenance rooms, respectively;
the actions to perform at respective regions are omitted and refer to~\cite{guo2016task} for methods to combine motion and action planning.
{To encourage visits over different office rooms and mimic the realistic scenario of victims
  being relocated from offices to medical stations,
  the label~\texttt{hm} is removed from an office after it has been visited certain number of times.
  For a more precise modeling with robot actions and dynamic environments, readers are referred to our earlier
  work~\cite{guo2015multi, guo2016task}.}
On the other hand, the safe-return constraints require the robot to be able to return to the base station ($\texttt{bs}$) from any medical station, via one designated exit ($\texttt{ex}$), i.e.,
\begin{equation}\label{eq:serach-and-rescue-safe}
\varphi_{\texttt{r}}= \Diamond (\texttt{md} \wedge\, \Diamond \texttt{ex}) \wedge \Diamond \Box \,(\bigvee\nolimits_{j\in J} \texttt{bs}_j),
\end{equation}
where~$J$ is the set of three base stations as shown in Fig.~\ref{fig:office-traj}.
For probabilistic requirements,
the satisfiability bound $\chi_{\texttt{o}}$ is set to $0.8$, while the safety bound $\chi_{\texttt{r}}$ is set to $0.9$.

\begin{figure}[t]
  \begin{center}
\begin{minipage}[t]{0.9\linewidth}
\centering
   \includegraphics[width =1.02\textwidth]{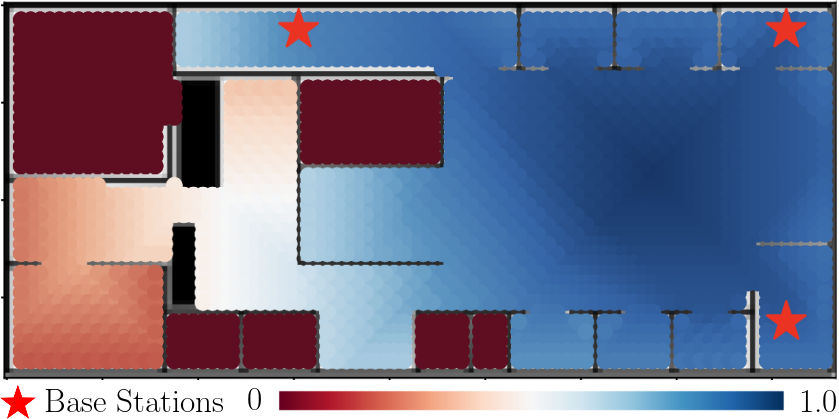}
  \end{minipage}
\end{center}
  \caption{Heatmap of the value function $v'^\star_{\texttt{r}}$ associated with the safe-return constraints in~\eqref{eq:serach-and-rescue-safe}.
The base station is marked by the red star.}
\label{fig:v-plot}
\end{figure}

\subsubsection{Simulation Results}\label{subsubsec:search-and-rescue-results}

In the section, we mainly present the results for the hierarchical planning algorithm in Alg.~\ref{alg:all}.
Comparison with other baselines are given in the sequel.
The discretization step for constructing the underlying MDP~$\mathcal{M}$ is set to $2\si{\meter}$ here,
which has $1100$ states and $10296$ edges.
\begin{table*}[t]
\begin{center}
  \resizebox{0.9\textwidth}{!}{
\begin{threeparttable}
	\caption{Scalability Results of Three Methods for the Case Study One.}
	\label{table:compare-search-and-rescue-complexity}
\begin{tabular}{ccccccccccccccc}
\toprule
\multirow{2}{*}{$\mathcal{M}$} & \multirow{2}{*}{Method\tnote{1}} & \multicolumn{2}{c}{{$\widehat{\mathcal{M}}$}, $\mathcal{M}$, $\mathcal{M}'$} & \multicolumn{2}{c}{$\mathcal{P}_{\texttt{r}}$, $\mathcal{P}'_{\texttt{r}}$} &
\multicolumn{2}{c}{{$\widehat{\mathcal{P}}$}, $\mathcal{P}_{\texttt{o}}$, $\mathcal{P}'_{\texttt{o}}$} &  \multicolumn{3}{c}{{$\widehat{\boldsymbol{\pi}}$}, $(\boldsymbol{\pi}_{\texttt{r}}, \boldsymbol{\pi}_{\texttt{o}})$, $(\boldsymbol{\pi}'_{\texttt{r}}, \boldsymbol{\pi}'_{\texttt{o}})$}\\
\cmidrule(lr){3-4}\cmidrule(lr){5-6}\cmidrule(lr){7-8}\cmidrule(lr){9-11}
& & Size\tnote{2} & {Time\tnote{4} [\si{\second}] } &
Size\tnote{2} & {Time [\si{\second}] } &
Size\tnote{2} & {Time [\si{\second}] } &
Size\tnote{3} & {Time [\si{\second}] } & Safety \\
\midrule
\multirow{ 3}{*}{(3\texttt{e}2,\;2\texttt{e}3)\tnote{2}} & {ALT} & {(6\texttt{e}2,\;6\texttt{e}3)} & {0.2} & {---/---} & {---/---} & {(2\texttt{e}5,\;2\texttt{e}7)} & {3\texttt{e}2} & {1\texttt{e}6} & {29.6} & {0.95}\\
\addlinespace[0.3em]
 & BSL & (3\texttt{e}2,\;2\texttt{e}3) &  \textbf{0.01} & (2\texttt{e}3,\;3\texttt{e}4) & 0.2 & (1\texttt{e}4,\;3\texttt{e}5) & 21.8 & 6\texttt{e}3 & 9.6 & 0.95\\
\addlinespace[0.3em]
 & HIR & (3\texttt{e}2,\;2\texttt{e}3) &  0.3 & (5\texttt{e}2,\;3\texttt{e}4) & \textbf{0.4} & (2\texttt{e}3,\;2\texttt{e}5) & \textbf{10.1} & 8\texttt{e}3 & \textbf{6.2} & 0.94\\
\midrule
\addlinespace[0.3em]
\multirow{ 3}{*}{(5\texttt{e}3,\;4\texttt{e}4)} & {ALT} & {(1\texttt{e}4,\;2\texttt{e}5)} & {0.5} & {---/---} & {---/---} & {(1\texttt{e}6,\;2\texttt{e}7)} & {4\texttt{e}3} & {8\texttt{e}7} & {2\texttt{e}4} & {0.9}\\
\addlinespace[0.3em]
& BSL & (5\texttt{e}3,\;4\texttt{e}4) & \textbf{0.3} & (4\texttt{e}4,\;3\texttt{e}5) & 3.6 & (2\texttt{e}5,\;2\texttt{e}6) & 165 & 4\texttt{e}5 & 3\texttt{e}3 & 0.9\\
\addlinespace[0.3em]
 & HIR & (3\texttt{e}2,\;4\texttt{e}3) & 8.3 & (5\texttt{e}2,\;4\texttt{e}4) & \textbf{2.3} & (4\texttt{e}3,\;2\texttt{e}5) & \textbf{18.7} & 5\texttt{e}4 & \textbf{7.1} & 0.91\\
\midrule
\addlinespace[0.3em]
\multirow{ 3}{*}{(2\texttt{e}4,\;2\texttt{e}5)} & {ALT} & {(4\texttt{e}4,\;6\texttt{e}5)} & {3.4} & {---/---} & {---/---} & {(9\texttt{e}6,\;5\texttt{e}7)} & {1\texttt{e}4} & {N/A\tnote{4}} & {N/A} & {N/A}\\
\addlinespace[0.3em]
 & BSL & (2\texttt{e}4,\;2\texttt{e}5) &  \textbf{1.4} & (1\texttt{e}5,\;2\texttt{e}6) & 15.8 & (8\texttt{e}5,\;7\texttt{e}6) & 7\texttt{e}2 & 2\texttt{e}6 & N/A & N/A\\
\addlinespace[0.3em]
 & HIR & (5\texttt{e}2,\;5\texttt{e}3) &  15.1 & (5\texttt{e}2,\;5\texttt{e}4) & \textbf{5.2} & (4\texttt{e}3,\;3\texttt{e}5) & \textbf{22.8} & 5\texttt{e}4 & \textbf{12.8} & 0.9\\
\midrule
\addlinespace[0.3em]
\multirow{ 3}{*}{(8\texttt{e}4,\;7\texttt{e}5)} & {ALT} & {(3\texttt{e}5,\;2\texttt{e}6)} &  {7.8} & {---/---} & {---/---} & {N/A\tnote{4}} & {N/A} & {N/A} & {N/A} & {N/A}\\
\addlinespace[0.3em]
 & BSL & (8\texttt{e}4,\;7\texttt{e}5) & \textbf{3.7} & (6\texttt{e}5,\;7\texttt{e}6) &61.2 & N/A & N/A & N/A & N/A & N/A\\
\addlinespace[0.3em]
 & HIR & (6\texttt{e}2,\;7\texttt{e}3) &  21.2 & (6\texttt{e}2,\;1\texttt{e}5) & \textbf{10.2} & (5\texttt{e}3,\;4\texttt{e}5) & \textbf{28.2} & 5\texttt{e}4 & \textbf{17.6} & 0.9\\
\bottomrule
\end{tabular}
  \begin{tablenotes}
  \item[1] {ALT: The alternative approach described in Sec.~\ref{subsec:alternative}
    and PRISM~4.6~\cite{kwiatkowska2011prism},
           which does not compute~$\mathcal{P}_{\texttt{r}}$ directly.
           BSL: The baseline solution in Alg.~\ref{alg:direct}.
           HIR: The hierarchical algorithm in Alg.~\ref{alg:all}.}
  \item[2] Sizes of the MDP models are measured by the number of nodes and edges.
    $a\texttt{e}b \triangleq a\times 10^b$. Note that $1\texttt{e}3\si{\second} \approx 16 \si{\minute}$
  \item[3] Sizes of the polices are measured by the number of LP variables.
  \item[4] {``N/A'' indicates either insufficient memory or computation time longer than $5$ hours}.
  \end{tablenotes}
\end{threeparttable}}
\end{center}
\end{table*}

First of all,
the DRA~$\mathcal{A}_{\texttt{r}}$ associated with $\varphi_{\texttt{r}}$ in~\eqref{eq:serach-and-rescue-safe}
is quite small with only $7$ states, $14$ edges and $1$ accepting pair.
Even though the original $\mathcal{M}$ is relatively large in size,
{the semi-MDP~$\mathcal{M}'_{\texttt{r}}$ in~\eqref{eq:abstract-mdp-safe} contains
$15$ states and $210$ edges}.
In particular, the synthesis of the low-level motion policy $\boldsymbol{\mu}'_{\texttt{o}}$
takes in average $3\si{\ms}$ for each edge within~$\mathcal{M}'_{\texttt{r}}$.
Afterwards, the product $\mathcal{P}'_{\texttt{r}}$ in~\eqref{eq:abstract-safe-product} is computed in $1.5\si{\second}$,
which has $511$ states and $36288$ edges.
Given $\mathcal{P}'_{\texttt{r}}$, the optimal return policy $\boldsymbol{\pi}'_{\texttt{r}}$, and its value function $v'^\star_{\texttt{r}}$ are computed in $2.3\si{\second}$.
A visualization of~$v'^\star_{\texttt{r}}$ is given in Fig.~\ref{fig:v-plot},
which matches our intuition well that states that can \emph{not} reach the base station have low values.
Specifically, due to the narrow passage between debris,
the left side of the office has relatively low value compared with the right side,
while the areas that are completely inaccessible have value close to zero.

Furthermore,
{the safety-ensured semi-MDP $\mathcal{M}'_{\texttt{o}}$ is computed}.
The DRA~$\mathcal{A}_{\texttt{o}}$ associated with task $\varphi$ in~\eqref{eq:serach-and-rescue-task} has~$39$ states, $255$ edges and $1$ accepting pair,
which is much larger than~$\mathcal{A}_{\texttt{r}}$ earlier.
It took $0.3\si{\second}$ to construct $\mathcal{M}'_{\texttt{o}}$ and
the resulting product~$\mathcal{P}'_{\texttt{o}}$.
At last, its associated outbound policy~$(\boldsymbol{\pi}'_{\texttt{o},\texttt{pre}}, \boldsymbol{\pi}'_{\texttt{o},\texttt{suf}})$ is computed in $5.1\si{\second}$ an $1.3\si{\second}$, respectively.
The complete computation time is listed in the second row of Table~\ref{table:compare-search-and-rescue-complexity}.
Given the outbound policy, Fig.~\ref{fig:office-traj} show several runs of the online execution.
It can be seen that the resulting trajectories search many different office areas and rescue the victims inside,
while avoiding such areas where the value function is low and thus not safe to return to the base station.
{To show how the safe-return constraints and task constraints affect the trajectories,
  the above procedure is repeated with different values of~$\chi_{\texttt{r}}$, $\chi_{\texttt{o}}$.
  More specifically, by setting these lower bounds to one of the values in~$\{0.0,0.5,0.9\}$,
  the resulting trajectories
  and the associated costs are summarized in Table~\ref{table:compare-different-bounds},
  as partially shown in Fig.~\ref{fig:office-traj}.
  {Note that the return event could be requested anytime after the system starts.}
  It can be seen that when~$\chi_{\texttt{o}}$ is small, the effectiveness of adding
  the safe-return constraints is not apparent as the robot simply stays around the initial state.
  However, when~$\chi_{\texttt{o}}$ is increased to improve task satisfiability and~$\chi_{\texttt{r}}$ increased to enforce the safety constraints,
  the resulting trajectories are more consistent and the robot is less likely to be trapped inside unsafe states.
  When~$\chi_{\texttt{r}}=0.0$, the average cost of the resulting trajectories is much higher with more variance,
  due to more trajectories where the robot is trapped at an early stage.
  Nonetheless, as shown in the last row of Table~\ref{table:compare-different-bounds},
  simply increasing~$\chi_{\texttt{o}}$ to~$0.9$ can \emph{not} achieve the same results.
  Namely, when~$\chi_{\texttt{o}}$ is set too high, the constrained optimization in~\eqref{eq:objective} becomes infeasible and thus no solutions exist.
  In that case, a relaxed and maximum-satisfying policy can be synthesized as proposed in our earlier work~\cite{guo2018probabilistic},
  which is outside the scope of this paper.
  One example trajectory when $\chi_{\texttt{r}}=0.0$ is shown in Fig.~\ref{fig:office-traj},
  which indicates that the policy can not distinguish the areas with different value functions,
  thus more prone to being trapped during execution.
  }

\begin{table}[t!]
  \begin{center}
 \resizebox{0.4\textwidth}{!}{
\begin{threeparttable}
    \caption{{Trajectory Cost under Different~$\chi_{\texttt{o}}$ and~$\chi_{\textup{\texttt{r}}}$.}}
	\label{table:compare-different-bounds}
\begin{tabular}{|c| c c c|}
 \hline
 {Traj. Cost\tnote{1}} & {$\chi_{\texttt{r}}=0.0$} & {$0.5$} & {$0.9$} \\ [0.5ex]
 \hline
 {$\chi_{\texttt{o}}=0.0$} & {$3.0\pm 1.8$} & {$4.3\pm 1.5$} & {$5.8\pm 1.2$} \\
 \hline
 {$0.5$} & {$56.4\pm 21.2$} & {$48.8\pm 12.3$} & {$42.3\pm 5.7$} \\
 \hline
 {$0.9$} & {N/A\tnote{2}} & {N/A} & {N/A} \\
 \hline
\end{tabular}
\begin{tablenotes}
\item[1] {Mean cost with standard deviation evaluated over~$100$ simulated runs
  of the proposed method with different~$\chi_{\texttt{o}},\chi_{\texttt{r}}$.
  If the robot is trapped, the cost is computed as the maximum action cost
multiplied by the horizon.}
\item[2] {``N/A'' indicates that no solutions can be found.}
  \end{tablenotes}
\end{threeparttable}}
\end{center}
\end{table}

\subsection{\textbf{Study Two}: Planetary Exploration Mission}\label{subsec:mars}
For the second numerical study,
we consider a planetary exploration mission similar to the DLR SpaceBot Camp~\cite{droeschel2017continuous}.
The rover-like mobile manipulator needs to navigate to different areas
to look for objects of interests, assemble and store them,
while maintaining its battery level by charging often.

\begin{table*}[t]
\begin{center}
  \resizebox{0.9\textwidth}{!}{
\begin{threeparttable}
	\caption{Scalability Results of Three Methods for the Case Study Two.}
	\label{table:compare-mars-complexity}
\begin{tabular}{ccccccccccccccc}
\toprule
\multirow{2}{*}{$\mathcal{M}$} & \multirow{2}{*}{Method\tnote{1}} & \multicolumn{2}{c}{{$\widehat{\mathcal{M}}$}, $\mathcal{M}$, $\mathcal{M}'$} & \multicolumn{2}{c}{$\mathcal{P}_{\texttt{r}}$, $\mathcal{P}'_{\texttt{r}}$} &
\multicolumn{2}{c}{{$\widehat{\mathcal{P}}$}, $\mathcal{P}_{\texttt{o}}$, $\mathcal{P}'_{\texttt{o}}$} &
\multicolumn{3}{c}{{$\widehat{\boldsymbol{\pi}}$}, $(\boldsymbol{\pi}_{\texttt{r}}, \boldsymbol{\pi}_{\texttt{o}})$, $(\boldsymbol{\pi}'_{\texttt{r}}, \boldsymbol{\pi}'_{\texttt{o}})$}\\
\cmidrule(lr){3-4}\cmidrule(lr){5-6}\cmidrule(lr){7-8}\cmidrule(lr){9-11}
& & Size & {Time [\si{\second}] } &
Size & {Time [\si{\second}] } &
Size & {Time [\si{\second}] } &
Size & {Time [\si{\second}] } & Safety \\
\midrule
\multirow{ 3}{*}{(9\texttt{e}2,\; 8\texttt{e}3)} & {ALT} & {(2\texttt{e}3,\;2\texttt{e}4)} & {0.1} & {---/---} & {---/---} & {(7\texttt{e}4,\;1\texttt{e}6)} & {24.1} & {7\texttt{e}3} & {9.6} & {0.95}\\
\addlinespace[0.3em]
 & BSL & (9\texttt{e}2,\; 8\texttt{e}3) & \textbf{0.03} & (6\texttt{e}3,\;7\texttt{e}4) & 0.6 & (3\texttt{e}4,\;2\texttt{e}5) & 6.7 & 2\texttt{e}4 & 6.7 & 0.93\\
\addlinespace[0.3em]
 & HIR & (1\texttt{e}2,\;2\texttt{e}3) & 0.2 & (1\texttt{e}2,\;3\texttt{e}3) & \textbf{1.9} & (2\texttt{e}3,\;1\texttt{e}5) & \textbf{3.5} & 1\texttt{e}4 & \textbf{2.5} & 0.92\\
\midrule
\addlinespace[0.3em]
\multirow{ 3}{*}{(2\texttt{e}4,\;2\texttt{e}5)} & {ALT} & {(4\texttt{e}4,\;6\texttt{e}5)} & {3.3} & {---/---} & {---/---} & {(8\texttt{e}4,\;1\texttt{e}7)} & {2\texttt{e}3} & {N/A} & {N/A} & {N/A}\\
\addlinespace[0.3em]
& BSL & (2\texttt{e}4,\;2\texttt{e}5) & \textbf{0.6} & (1\texttt{e}5,\;1\texttt{e}6) & 10.8 & (4\texttt{e}5,\;4\texttt{e}6) & 145 & 3\texttt{e}5 & 2\texttt{e}3 & 0.9\\
\addlinespace[0.3em]
 & HIR & (2\texttt{e}2,\;1\texttt{e}4) & 8.3 & (4\texttt{e}2,\;4\texttt{e}4) & \textbf{2.9} & (2\texttt{e}3,\;1\texttt{e}5) & \textbf{8.7} & 2\texttt{e}4 & \textbf{4.1} & 0.9\\
\midrule
\addlinespace[0.3em]
\multirow{ 3}{*}{(6\texttt{e}4,\;7\texttt{e}5)} & {ALT} & {(2\texttt{e}4,\;2\texttt{e}6)} & {10.4} & {---/---} & {---/---} & {N/A} & {N/A} & {N/A} & {N/A} & {N/A}\\
\addlinespace[0.3em]
 & BSL & (6\texttt{e}4,\;7\texttt{e}5) & \textbf{2.4} & (4\texttt{e}5,\;5\texttt{e}6) & 50.4 & (2\texttt{e}6,\;2\texttt{e}7) & 4\texttt{e}2 & N/A & N/A & N/A\\
\addlinespace[0.3em]
 & HIR & (3\texttt{e}2,\;2\texttt{e}4) & 19.1 & (5\texttt{e}2,\;4\texttt{e}4) & \textbf{3.2} & (2\texttt{e}3,\;2\texttt{e}5) & \textbf{10.1} & 2\texttt{e}4 & \textbf{6.3} & 0.9\\
\midrule
\addlinespace[0.3em]
\multirow{ 3}{*}{(3\texttt{e}5,\;3\texttt{e}6)} & {ALT} & {(6\texttt{e}5,\;1\texttt{e}7)} &  {1\texttt{e}2} & {---/---} & {---/---} & {N/A} & {N/A} & {N/A} & {N/A} & {N/A}\\
\addlinespace[0.3em]
 & BSL & (3\texttt{e}5,\;3\texttt{e}6) & \textbf{18.7} & N/A & N/A & N/A & N/A & N/A & N/A & N/A\\
\addlinespace[0.3em]
 & HIR & (4\texttt{e}2,\;4\texttt{e}4) & 25.1 & (5\texttt{e}2,\;5\texttt{e}4) & \textbf{8.4} & (2\texttt{e}3,\;3\texttt{e}5) & \textbf{18.2} & 3\texttt{e}4 & \textbf{10.6} & 0.9\\
\bottomrule
\end{tabular}
  \begin{tablenotes}
  \item[1] {Legends are the same as in Table~\ref{table:compare-search-and-rescue-complexity}.}
  \end{tablenotes}
\end{threeparttable}}
\end{center}
\end{table*}

\subsubsection{Workspace and Task Description}\label{subsubsec:mars-ws}
As shown in Fig.~\ref{fig:island-traj}, a rover of size $1m\times 1m$ is deployed to a planetary-like environment of size $90m\times 90m$ with rough terrains.
Its dynamic model within the environment is similar to the UGV described in Sec.~\ref{subsubsec:search-and-rescue-ws}.
States are marked by objects of interests with different probabilities.
The environment consists of different types of soil surfaces that may cause slip and even trapping.
Furthermore, there are also hills and valleys that might be too steep to descend and ascend.
Once trapped, it may not be able to return safely to its base station.
{The system model is initialized by a depth image with a desired discretization level
  and manual estimation of the labels.
Namely, given this depth image and the robot motion model above,
  the system model~$\mathcal{M}$ can be constructed algorithmically,
  by checking the relative depth between neighboring cells.}

\begin{figure}[t]
  \begin{center}
    \includegraphics[width =0.95\linewidth]{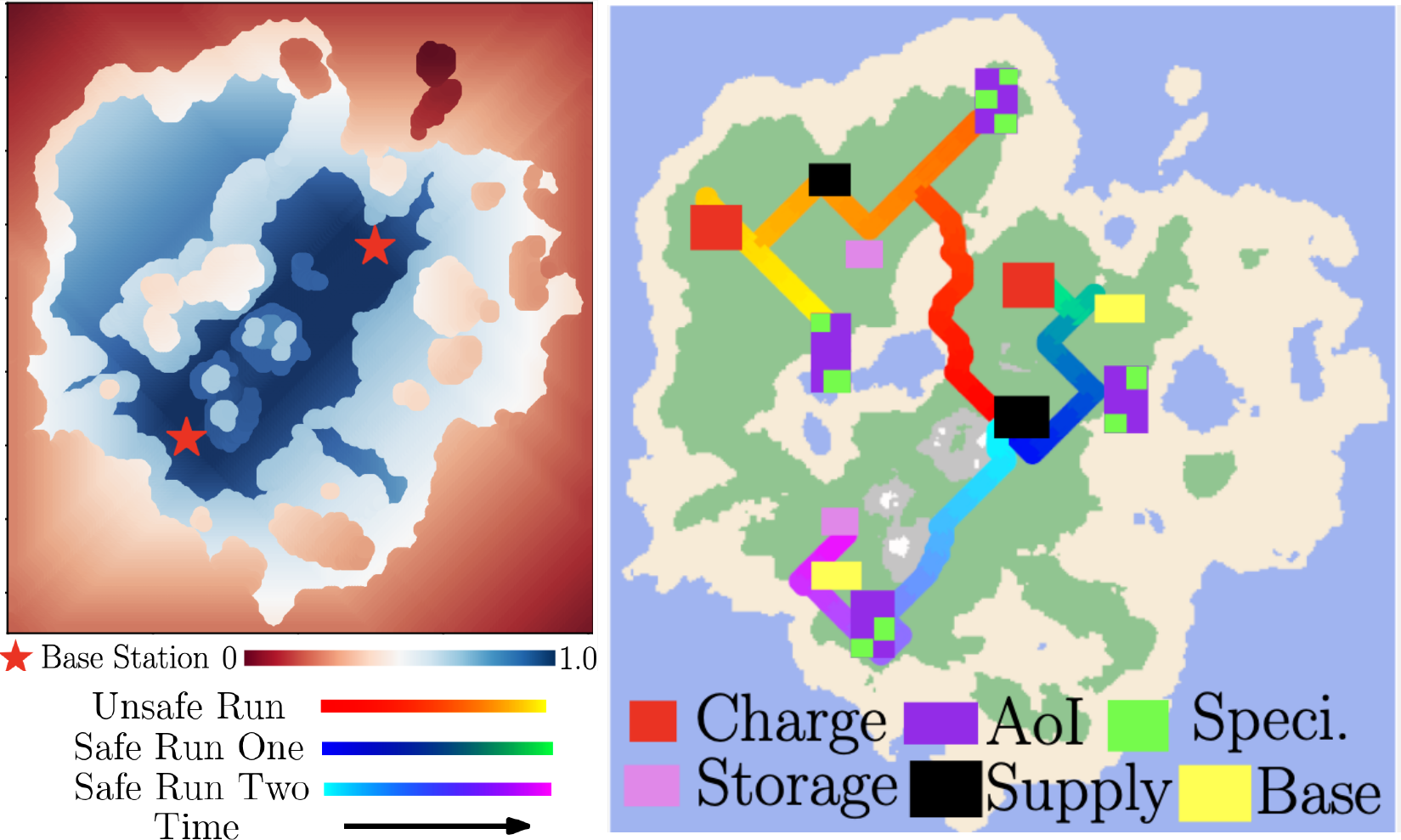}
  \end{center}
  \vspace{-0.2in}
  \caption{\textbf{Left}: heatmap of the value function $v'^\star_{\texttt{r}}$ associated with the safe-return constraints in~\eqref{eq:mars-safe}.
The base station is marked by the red star;
\textbf{Right}: examples of different runs under the outbound policy~$\boldsymbol{\mu}'_{\texttt{o}}$ and an unsafe policy (in red) without the safe constrains.}
\label{fig:island-traj}
\end{figure}


More specifically, the exploratory mission include:
(i) navigate to several potential areas of interest ($\texttt{ar}_i$) and scope specimens ($\texttt{ss}$);
(ii) navigate to a supply area ($\texttt{sp}$), grasp containers and put specimens in the containers;
(iii) navigate to a storage area ($\texttt{st}$) and put the containers there;
finally (iv) charge at the charging areas ($\texttt{ch}$) whenever battery is low.
These tasks can be specified as the following LTL formulas:
\begin{equation}\label{eq:mars-task}
 \begin{split}
   \varphi =\; &(\Box \Diamond \texttt{ch})  \wedge \Big(\Box \Diamond (\bigvee\nolimits_{i\in I} \texttt{ar}_i \wedge \texttt{ss})  \\
   & \wedge \Box (\texttt{ss} \rightarrow (\neg \texttt{ss})\, \textsf{U}\, \Diamond (\texttt{sp}\wedge \Diamond \texttt{st})) \Big),
   \end{split}
\end{equation}
where $I$ is the set of potential areas with specimens.
{Similar to the previous case,
  to encourage visits over different areas of interest,
  the label~\texttt{ss} is removed from an area after it has been visited a certain number of times.
  The action model and the method to dynamically update the environment model are omitted and
  refer to our earlier work~\cite{guo2015multi, guo2016task}.}
On the other hand, the safe-return constraints require the rover to be able to
return to the base station ($\texttt{bs}$) after charging ($\texttt{ch}$) and visiting the two regions, i.e.,
\begin{equation}\label{eq:mars-safe}
  \varphi_{\texttt{r}}= \Diamond (\texttt{ch} \wedge \Diamond (\bigvee\nolimits_{i\in I}\texttt{ar}_i)) \wedge \Diamond\Box\, (\bigvee\nolimits_{j\in J}\texttt{bs}_j).
\end{equation}
where $J$ is the set of two base stations as shown in Fig.~\ref{fig:island-traj}.
Due to high uncertainties in the model, the satisfiability bound~$\chi_{\texttt{o}}$ is set to $0.6$,
while the safety bound~$\chi_{\texttt{r}}$ is set to $0.9$.

\subsubsection{Simulation Results}\label{subsubsec:mars-results}
The results by following Alg.~\ref{alg:all} are first presented here.
By setting the discretization step to~$3\si{\meter}$,
the underlying MDP~$\mathcal{M}$ has $900$ states and $8416$ edges.
Note that this model is highly \emph{non-ergodic},
meaning that many states can be reached \emph{from} the base station,
but can not reach the base station.
For instance,
as shown in Fig.~\ref{fig:island-traj}, some valleys can be reached easily by descending,
but impossible to ascend back.
First, given the DRA~$\mathcal{A}_{\texttt{r}}$ associated with~$\varphi_{\texttt{r}}$ in~\eqref{eq:mars-safe},
{the semi-MDP~$\mathcal{M}'_{\texttt{r}}$ and its motion policy~$\boldsymbol{\mu}'_{\texttt{r}}$ are constructed in~$1.7\si{\second}$}.
The resulting product~$\mathcal{P}'_{\texttt{r}}$ has $119$ nodes and $2628$ edges,
of which the optimal return policy $\boldsymbol{\pi}'_{\texttt{r}}$,
and the value function $v'^\star_{\texttt{r}}$ are computed in $1.9\si{\second}$.
The distribution of~$v'^\star_{\texttt{r}}$ is shown in Fig.~\ref{fig:island-traj}.
It can be noticed that the probability of returning to the base station decreases each time a valley or a rough terrain is crossed.

Second, the task DRA~$\mathcal{A}_{\texttt{o}}$ has~$26$ states, $190$ edges and~$1$ accepting pair,
which is slightly simpler than the search-and-rescue task.
{The semi-MDP~$\mathcal{M}'_{\texttt{o}}$ and its motion policy $\boldsymbol{\mu}'_{\texttt{o}}$ are computed in $1.3\si{\second}$ with $37$ states and $1332$ edges}.
Then, their product~$\mathcal{P}'_{\texttt{o}}$ is constructed in~$3.5\si{\second}$, with $1898$ states and $134784$ edges.
At last, its associated outbound policy~$\boldsymbol{\pi}'_{\texttt{o}}$ is computed in $2.5\si{\second}$ via a LP of $10^4$ variables.
The detailed model size and computation time are reported in the second row of Table~\ref{table:compare-mars-complexity}.
Fig.~\ref{fig:island-traj} shows several trajectories by following the hierarchical task policy during online execution.
It can be seen that they remain mostly within central plain area, where the value function is high.
In contrast,
the above procedure is repeated without the safe-return constraints.
One such run is also shown in Fig.~\ref{fig:island-traj},
which instead crosses the connecting valleys to reach the upper-left region where the value function is low,
and thus more likely to get trapped and not be able to return the base station.

\begin{figure}[t]
  \begin{center}
\begin{minipage}[t]{0.99\linewidth}
\centering
\includegraphics[width=0.49\textwidth]{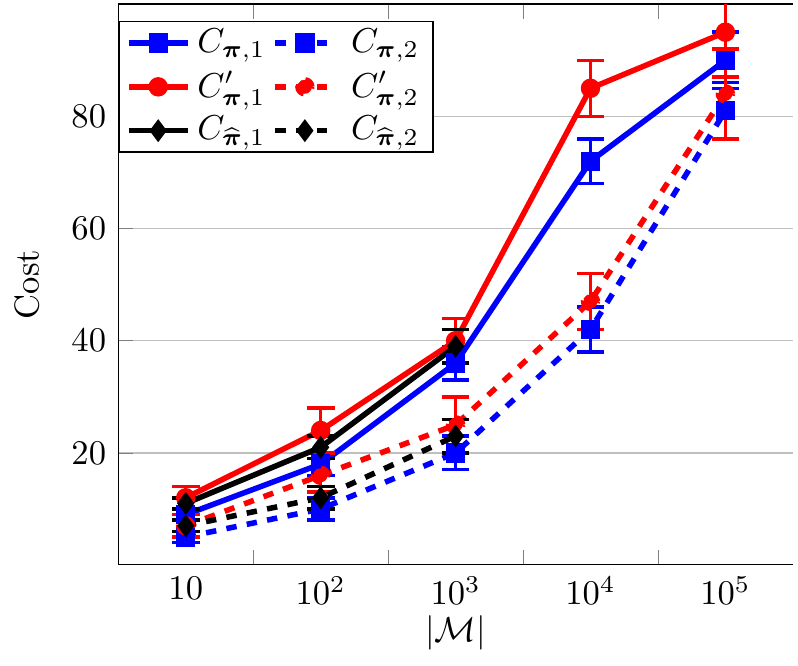}
\includegraphics[width=0.49\textwidth]{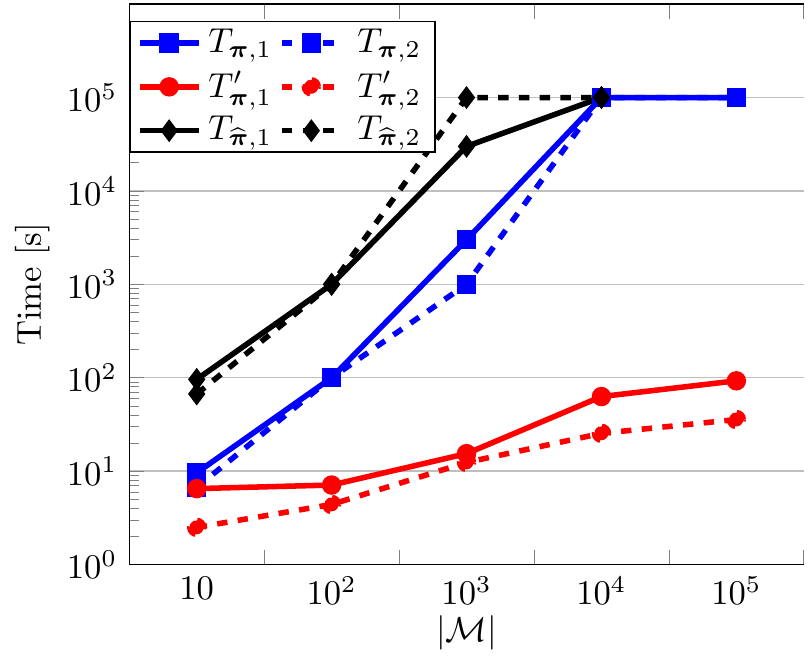}
\end{minipage}
  \end{center}
  \vspace{-0.2in}
  \caption{
    \textbf{Left}: Cost distribution of~$100$ runs under the polices generated by the baseline solution $(C_{\boldsymbol{\pi},1},~C_{\boldsymbol{\pi},2}$),the hierarchical algorithm $(C'_{\boldsymbol{\pi},1},~C'_{\boldsymbol{\pi},2})$
    {and the alternative approach~$(C_{\widehat{\boldsymbol{\pi}},1}$, $C'_{\widehat{\boldsymbol{\pi}},2})$.}
    \textbf{Right}: Evolution of solution time for both case studies w.r.t. the system size $|\mathcal{M}|$:
    $(T_{\boldsymbol{\pi},1}, T_{\boldsymbol{\pi},2})$ for the baseline solution,
    $(T'_{\boldsymbol{\pi},1}, T'_{\boldsymbol{\pi},2})$ for the hierarchical algorithm,
    {and $(T_{\widehat{\boldsymbol{\pi}},1}, T_{\widehat{\boldsymbol{\pi}},2})$ for the alternative approach.}}
\label{fig:compare}
\end{figure}

\subsection{Comparison with Baselines}\label{subsubsec:exp-compare}

To further validate the computational gain and cost optimality of the proposed hierarchical approach (HIR),
{we compare it with the baseline method (BSL) described in Alg.~\ref{alg:direct}
  and the alternative method (ALT) discussed in Sec.~\ref{subsec:alternative} for both case studies.
  Regarding the ALT method, PRISM~4.6 is used~\cite{kwiatkowska2011prism} with the ``multi-objective solution method'' option set to~``LP'', see~\cite{Forejt12Pareto} to compute the prefix policies,
  while a similar algorithm as proposed in our earlier work~\cite{guo2018probabilistic} is followed
  to compute the suffix policy within the AMECs.
}
More specifically, we decrease the discretization size of the workspace such that
the size of~$\mathcal{M}$ increases gradually.

{For the case study one, Table~\ref{table:compare-search-and-rescue-complexity} summarizes the resulting size of $\mathcal{M}_{\texttt{r}}$, $\mathcal{M}_{\texttt{o}}$, $\mathcal{P}_{\texttt{r}}$ and $\mathcal{P}_{\texttt{o}}$ for Alg.~\ref{alg:direct};
the size of~$\widehat{\mathcal{M}}$,~$\widehat{\mathcal{P}}$ for the alternative method;
and the size of $\mathcal{M}'_{\texttt{r}}$, $\mathcal{M}'_{\texttt{o}}$, $\mathcal{P}'_{\texttt{r}}$ and $\mathcal{P}'_{\texttt{o}}$ for Alg.~\ref{alg:all}.
Note that since the alternative method computes directly the extended model~$\widehat{\mathcal{M}}$ and its product~$\widehat{\mathcal{P}}$,
the computation of product~$\mathcal{P}_{\texttt{r}}$ does not apply}.
The computation time for these models and the associated polices are also reported.
{
Both the alternative method and the baseline method take less pre-processing time to compute~$\widehat{\mathcal{M}}$ and~$\mathcal{M}$, compared with the abstracted model~$\mathcal{M}'$.
However, as the system size increases, it is clear that both methods quickly become intractable and even formulating the underlying LPs over~$\mathcal{P}_{\texttt{o}}$ takes hours.
For MDPs with more than~$0.7$ million edges,
the alternative method fails to generate the product~$\widehat{\mathcal{P}}$ before even computing the overall policy~$\widehat{\boldsymbol{\pi}}$.
This is mainly due to the fact that~$\widehat{\mathcal{P}}=\widehat{\mathcal{M}}\times \mathcal{A}_{\texttt{r}} \times \mathcal{A}_{\texttt{o}}$,
which leads to a drastic blow-up of the model size,
compared with~$\mathcal{P}_{\texttt{o}}$ or~$\mathcal{P}'_{\texttt{o}}$ in the other two methods.
Although the baseline method can compute the outbound and safety products~$\mathcal{P}_{\texttt{r}},\,\mathcal{P}_{\texttt{o}}$,
neither the return policy nor the outbound policy can be computed within reasonable amount of time, i.e.,~$5$ hours.}
In contrast, the proposed method not only can solve the same set of problems with~at least~$10$ times less time,
but also problems with much higher complexity where the baseline method failed.
It is interesting to notice that the computation of the abstraction model $\mathcal{M}'_{\texttt{r}}$, $\mathcal{M}'_{\texttt{o}}$ and the associated motion policy took most of the time,
while the size of the product model $\mathcal{P}'_{\texttt{r}}$ and $\mathcal{P}'_{\texttt{o}}$ remains relatively constant given a fixed task specification.

Similar analyses are performed for the case study two.
Table~\ref{table:compare-mars-complexity} reports similar results.
{Namely,
  both the alternative method and the baseline method fail to generate a solution
  for systems with a large number of states and edges,
  while the proposed hierarchical approach is close to one order of magnitude more efficient.
For the extreme case where~$\mathcal{M}$ has~$3$ million edges,
the alternative approach or the baseline solution can neither construct the products nor the linear programs given the memory and time limits,
while the proposed approach can obtain the task and motion policy within~$37.2\si{\second}$.}
Fig.~\ref{fig:compare} plots the solution time with respect to the system size for both cases.

Last but not least, the cost optimality of the resulting polices {from all three methods} are evaluated by $100$ Monte Carlo simulations, for both case studies.
As shown in Fig.~\ref{fig:compare},
{for system sizes where the baseline method can find the optimal solution,
  the proposed approach has a close-to-optimal cost,
while the alternative method can only solve quite limited scenarios.}
Even for the case where $\mathcal{M}$ has around~$10^5$ edges, the proposed approach is within the range of $5\%$ extra cost.

\subsection{Hardware Experiment}\label{subsec:hardware}

The hardware experiment is carried out on an autonomous ground vehicle,
within an artificial office environment, as shown in Fig.~\ref{fig:lab_snaps}.
The workspace has a size of~$5 \si{\meter}\times 4 \si{\meter}$.
Its state is monitored by a motion capture system and the communication between
the state estimation, planning and control is handled by Robot
Operating System (ROS).

\begin{figure}[t]
  \begin{center}
\begin{minipage}[t]{0.99\linewidth}
\centering
\includegraphics[height=0.49\textwidth]{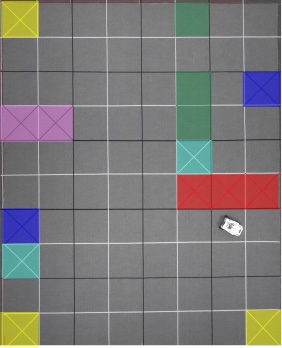}
\includegraphics[height=0.49\textwidth]{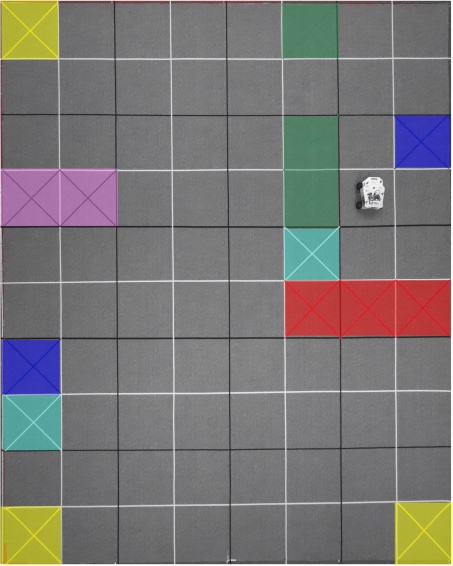}
\end{minipage}
  \end{center}
  \vspace{-0.2in}
  \caption{Snapshots of the task execution.
    \textbf{Left}: the robot is transporting victims from one
    \texttt{bs} to one \texttt{md}.
    \textbf{Right}: without the safe-return constraint, the robot reaches
    the other \texttt{md} after crossing the narrow passage between \texttt{str}.}
  \label{fig:lab_snaps}
\end{figure}

\subsubsection{Robot and Workspace Description}\label{subsubsec:hw-workspace}
As shown in Fig.~\ref{fig:lab_snaps},
the workspace is divided into~$10 \times 8$ cells.
The features and task description
follow a similar setting to the search and rescue mission described in
Sec.~\ref{subsubsec:search-and-rescue-ws}.
Similar to~\eqref{eq:serach-and-rescue-task},
the task is to transport victims in base stations
($\texttt{bs}_1$,\, $\texttt{bs}_2$) to any medical stations~$\texttt{md}$,
while avoiding the obstacles.
On the other hand, the safe-return constraint is similar
to~\eqref{eq:serach-and-rescue-safe},
which requires the vehicle to return to the first base~$\texttt{bs}_3$,
without being trapped in the stairs.
The number of offices and obstacles is smaller than the simulated case,
of which the associated probabilities are shown in Fig.~\ref{fig:compare-trajs}.
Note that the satisfiability bound~$\chi_\texttt{o}$ is set to~$0.75$
and the safe-return bound~$\chi_{\texttt{r}}$
is set to~$0.95$.
The vehicle has a navigation controller to move from any cell to adjacent cell
in a ``turn-and-forward'' fashion.
It results in a similar non-determinism, i.e.,
it drifts side-ways while moving and overshoot while rotating,
but with a smaller uncertainty compared to the simulated cases.
{Then, the complete model~$\mathcal{M}$ is constructed automatically by composing
this motion model with the labeled workspace model.}

\subsubsection{Results}\label{subsubsec:results}
In this part, we first report the results obtained by the baseline method
in comparison with the case where no safe-return constraint is imposed.
Then we show that the proposed hierarchical algorithm can generate
the same safe-policy but with only a fraction of the planning time.

First, via the baseline method, the product~$\mathcal{P}_{\texttt{o}}$ has $9880$
states and $87048$ edges, while the~$\mathcal{P}_{\texttt{r}}$ has~$1900$
states and $16740$ edges.
It takes~$30.8\si{\second}$ and~$0.1\si{\second}$  to compute the
policy~$\boldsymbol{\pi}_{\texttt{r}}$ and~$\boldsymbol{\pi}_{\texttt{o}}$,
respectively.
One of the resulting trajectories is shown in Fig.~\ref{fig:compare-trajs},
which avoids the regions which can only be accessed via stairs.
In comparison, when no such safe-return requirements as
in~\eqref{eq:serach-and-rescue-safe} are imposed,
the same product automaton~$\mathcal{P}_{\texttt{o}}$ can be used and one resulting
trajectory is also shown in Fig.~\ref{fig:compare-trajs}.
It can be seen that the robot reaches the medical station~\texttt{md}
via the narrow passage
is trapped when exiting one region.
Moreover, the proposed hierarchical algorithm is applied to the same
problem.
{It took around~$2.1\si{\second}$ to construct the
semi-MPDs~$\mathcal{M}_{\texttt{o}}'$ and~$0.5\si{\second}$
for~$\mathcal{M}_{\texttt{r}}'$}.
Afterwards, the high-level task policy~$\boldsymbol{\pi}'_{\texttt{o}}$
is synthesized in~$7.3\si{\second}$, while the safe policy and
the value function~$v'^{\star}$ is computed in~$2.6\si{\second}$.
The total planning time is around half of the baseline solution.
However, if the workspace is {expanded} by \emph{four} times the size,
the baseline solution takes around~$97.5\si{\second}$ while
the hierarchical solution generates the optimal polices in
merely~$10.7\si{\second}$,
i.e., close to one order of magnitude reduction in planning time,
which is similar to the trend observed in the simulated cases.

\begin{figure}[t]
  \begin{center}
\begin{minipage}[t]{0.99\linewidth}
\centering
\includegraphics[width=0.49\textwidth]{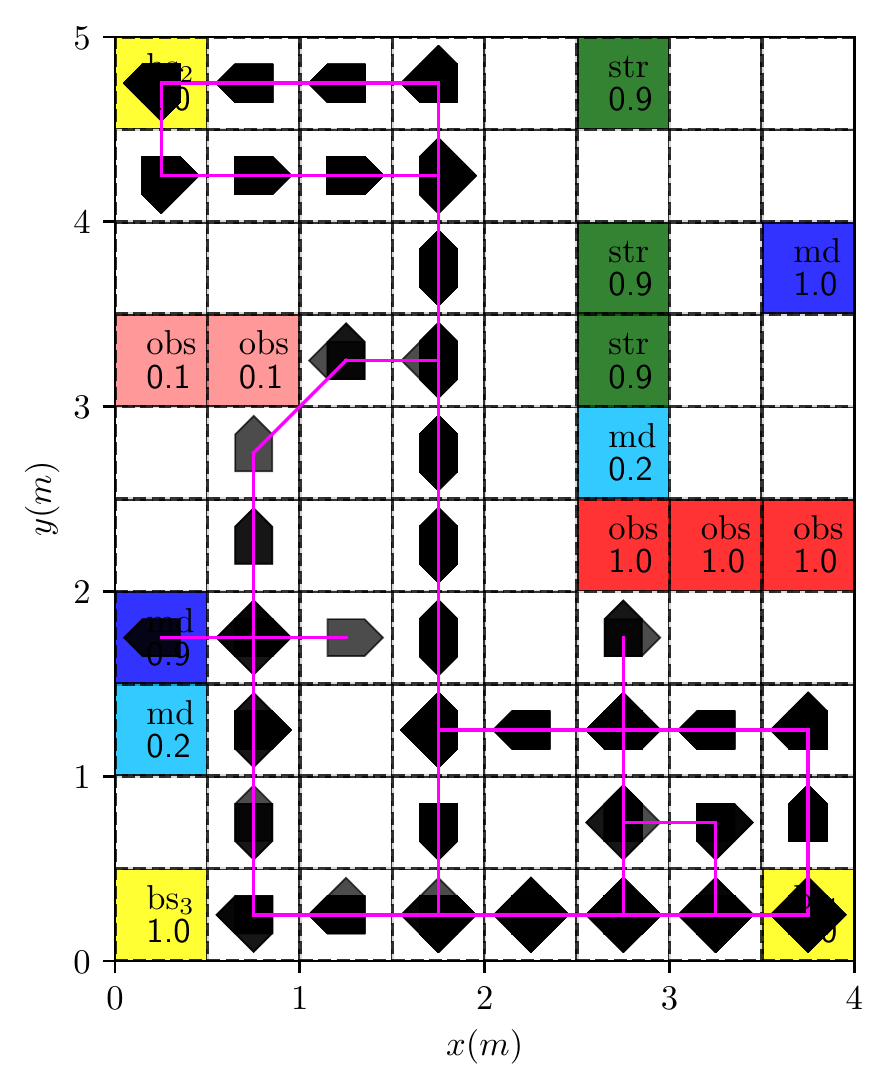}
\includegraphics[width=0.49\textwidth]{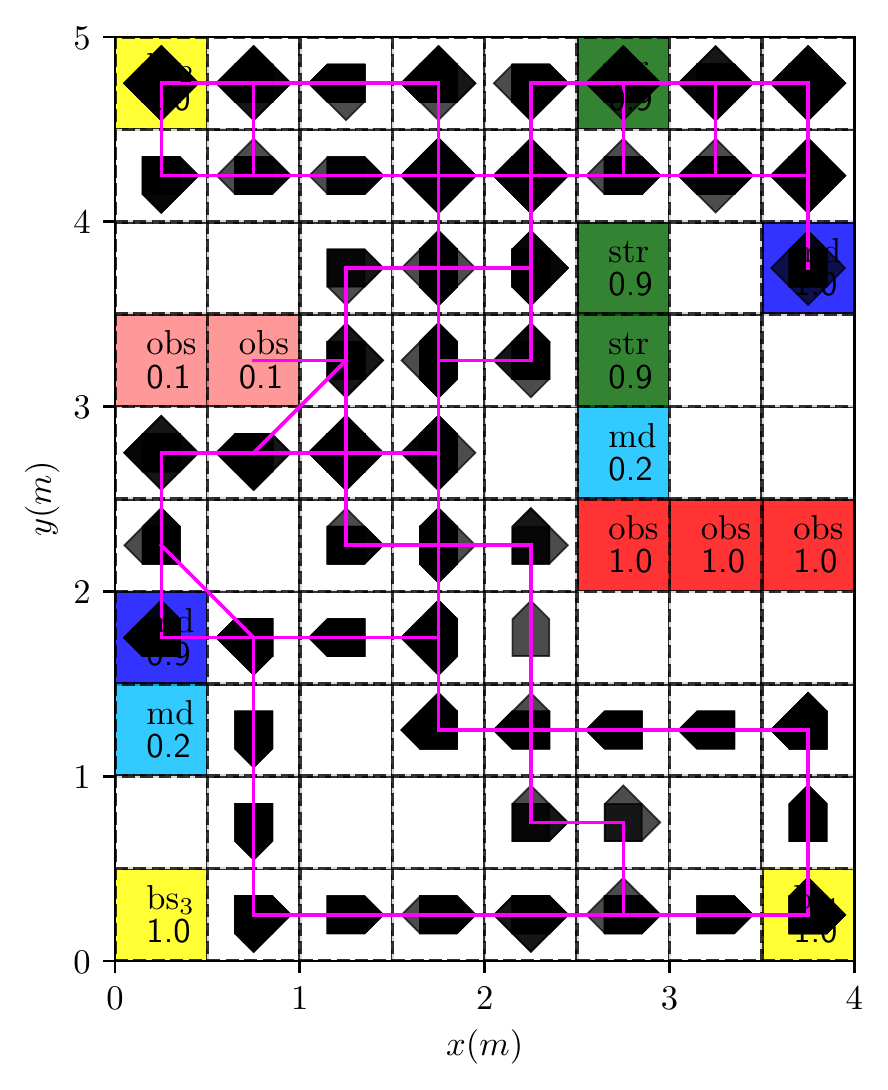}
\end{minipage}
  \end{center}
  \vspace{-0.2in}
  \caption{Examples of robot trajectories in the plan suffix
    with (\textbf{left}) and without (\textbf{right})
    the safe-return constraints via the proposed approach.
}
\label{fig:compare-trajs}
\end{figure}

\section{Summary and Future Work}\label{sec:future}
This work has proposed a hierarchical motion planning algorithm for mobile robots operating within uncertain environments.
The proposed algorithm has taken into account not only high-level tasks but also safe-return constrains, both of which are specified as LTL formulas.
It has been shown that the hierarchical planning algorithm significantly reduces the computation time compared with baseline solution, while maintaining a close-to-optimal performance.
Future work includes online adaptation of the hierarchical policies.


\bibliographystyle{IEEEtran}
\bibliography{main}


\end{document}